\documentclass[10pt,twocolumn,letterpaper]{article}

\usepackage{cvpr}
\usepackage{times}
\usepackage{epsfig}
\usepackage{graphicx}
\usepackage{amsmath}
\usepackage{soul}
\usepackage{amssymb}
\usepackage[noend]{algpseudocode}
\usepackage{color, colortbl}
\usepackage{makecell}

\definecolor{ColorRow}{rgb}{1,1,1}
\definecolor{ColorTable}{rgb}{0.9,0.9,0.9}

\usepackage[utf8]{inputenc}
\usepackage[english]{babel}
\usepackage{amsfonts}
\usepackage{caption}
\usepackage{multirow}
\usepackage{stackengine}

\usepackage{xcolor}
\usepackage{microtype}
\usepackage{subcaption}
\usepackage{algorithm2e}

\usepackage[breaklinks=true,bookmarks=false]{hyperref}

\cvprfinalcopy 

\def\cvprPaperID{7769} 

\ifcvprfinal\fi
\begin{document}

\setlength{\abovedisplayskip}{5pt}
\setlength{\belowdisplayskip}{5pt}

\title{Optimal least-squares solution to the hand-eye calibration problem}


\author{Amit Dekel \quad Linus H\"{a}renstam-Nielsen \quad Sergio Caccamo\\
	Univrses\\
	{\tt\small \{amit.dekel, linus.arenstam, sergio.caccamo\}@univrses.com}
}

\maketitle


\begin{abstract}
We propose a least-squares formulation to the noisy hand-eye calibration problem using dual-quaternions,
and introduce efficient algorithms to find the exact optimal solution, based on analytic properties of the problem, avoiding non-linear optimization.
We further present simple analytic approximate solutions which provide remarkably good estimations compared to the exact solution.
In addition, we show how to generalize our solution to account for a given extrinsic prior in the cost function.
To the best of our knowledge our algorithm is the most efficient approach to optimally solve the hand-eye calibration problem.

\end{abstract}


\section{Introduction}
Hand-eye calibration is a common problem in computer vision where one wishes to find the transformation between two rigidly attached frames. One popular formulation of the problem is in terms of the following equation
 \begin{align}\label{eq:CXXO}
 \Delta C \circ X = X \circ \Delta H,
 \end{align}
 where $\Delta C, \Delta H, X$ are $SE(3)$ elements and $\circ$ is the group multiplication operation \cite{shiu1989calibration}.
 $\Delta C$ and $\Delta H$ represent the sources relative poses (\eg camera and robot hand) and $X$ is the rigid transformation between the two frames.
 
 In the literature one can find plenty of strategies to solve the problem for a given set of poses, as well as different representations and metrics being used, see \cite{Esquivel2015EyetoeyeCE, Shah:2012:ORC:2393091.2393095} for a recent review.
 Since the $SE(3)$ group is a semi-direct product of the special orthogonal group and the translational group ($SO(3) \ltimes T(3)$), starting with (\ref{eq:CXXO}), one usually ends up with two equations which we will refer to as the rotational equation (the quotient by the normal subgroup) and the translational equation (the normal subgroup). 
 In practice, due to noisy data these equations cannot be satisfied exactly, and a numerical minimization approach or an approximating procedures are often used.
 
 \begin{figure}[t]
  \begin{center}
   \includegraphics[width=0.84\linewidth]{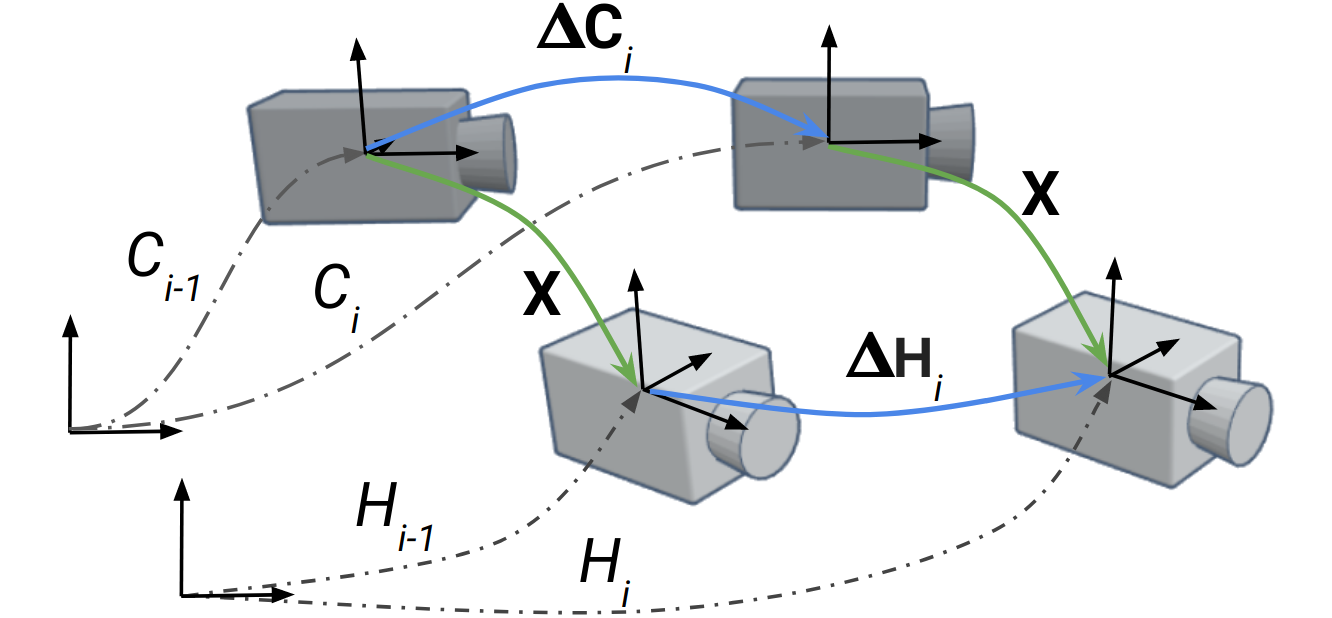}
   \caption{\small{The hand-eye calibration problem for two rigidly attached cameras.}}
   \label{fig:hand_eye_setup}
  \end{center}
\end{figure}


 Some authors approach the problem by solving the two equations separately by first minimizing a term related to the rotational equation and then substituting the solution in the translational equation and minimizing this term in turn, where a closed form solution can be found \cite{Tsai:1988:NTF:57425.57456, Park1994RobotSC, horaud1995hand}.
 Others try to minimize both terms simultaneously.

 In the later case, one approach is to perform a (constrained) nonlinear optimization procedure to minimize a cost function which takes into account the two terms (the rotational and translational parts) in some representation and specified metrics. This procedure is more costly in terms of computations, and requires a proper initial estimation in order to avoid local minima \cite{horaud1995hand, doi:10.1177/027836499101000305, Strobl2006OptimalHC}.
 A linear simultaneous solution was presented in terms of dual-quaternions (or equivalently geometric algebra) \cite{Daniilidis1999HandEyeCU}, \cite{Bayro-Corrochano2000}, which aims to solve (\ref{eq:CXXO}) rather than a minimization problem, being valid for the noise free case, or requires filtering of the data.
 
 As mentioned above, one can use different metrics for the minimization problem, usually related to the chosen representations or noise model assumptions. For the rotational part, the common (non-equivalent \cite{Hartley2013}) choices are the chordal, quaternionic or the geodesic metrics.
 Similarly, one can use different minimization terms for the translational part.
 Furthermore, rotation and translation terms have different units, and a proper relative weighting should be used. A natural choice would be the covariance of the measurements, which is not always available.
 
 In this paper we address the problem of simultaneous minimization using the dual-quaternions (DQs) representation.
 The difficulties in solving the minimization problems analytically are due to nonlinear constraints. Using DQs, one can construct a quadratic cost function in terms of the DQ \dof which are subject to nonlinear constraints, which inevitably make the problem highly nonlinear.
 
 Our approach is to add the constraints as two Lagrange multiplier terms and to study their analytic properties. 
 The two Lagrange multipliers are related by a polynomial $p(\lambda, \mu)$ of 8th-order in $\mu$ and 4th-order in $\lambda$, with some special properties.
 Over the reals, the polynomial defines four curves $\lambda_i(\mu)$, and the optimal solution corresponds to the saddle point $d\lambda / d\mu = 0$ of the smallest $\lambda$ curve.
 Thus, our strategy is to efficiently find this special point in the Lagrange multipliers space.
 We propose several nonequivalent ways to find the optimal solutions as well as a hierarchy of analytic approximations which perform remarkably well.
 
 We show explicitly how to extend our algorithms when adding a prior to the problem, allowing for a maximum a posteriori (MAP) analysis, or regularization in degenerate cases (\eg planar motion).

 We perform several experiments using synthetic and real data, and compare our algorithms to other available methods and to nonlinear optimization, showing that our algorithm indeed finds the optimal solution.
 
 The paper is organized as follows: we initially establish the connection between the $SE(3)$ group and unit DQ. Then we formulate the hand-eye calibration problem using DQs, and introduce the minimization problem. 
 Afterwards we discuss the properties of the cost function and present algorithms for solving the problem.
 Next, we extend the formulation to include a prior term.
 In the final sections, we present our experiments and comparisons to other existing methods in the literature, along with discussion and conclusions. 
 Some technical details and more detailed experiment results are relegated to appendices.

\section{SE(3) dual-quaternions representation}
\label{sec:DualQuaternion}
	 
In this paper we use the dual-quaternion representation of the $SE(3)$ group.
The group elements are constructed using dual-quaternions which are a sum of an ordinary quaternion and a dual part which is an ordinary quaternion multiplied by the dual-number $\epsilon$, where $\epsilon^2 = 0$, see for example \cite{Jia, Daniilidis1999HandEyeCU}. 
 
$SE(3)$ can be represented by unit DQ, $Q = q + \epsilon q'$ with $Q \otimes Q^* = 1$, where $q, q' \in \mathbb{H}$, $\otimes$ is the quaternionic product, and ${}^*$ is the usual quaternion conjugation. As with ordinary quaternion double cover representation of representing $SO(3)$, we identify $Q \sim -Q$. 
 The unit DQ constraint can be equivalently written as $|q| = 1$ and $q\cdot q' = 0$ where $q$ and $q'$ are treated as real four-dimensional vectors, and $\cdot$ is the dot product.
 
The explicit, a rotation element is given by $R_{(\hat k, \theta)} = (\hat k \sin\frac{\theta}{2}, \cos\frac{\theta}{2})$ with $\hat k$ the rotation axis and $\theta$ the rotation angle, while a translation element by a vector $\vec a$ is given by $T_{\vec a} = (\vec 0, 1) + \epsilon \frac{1}{2}(\vec a, 0)$. 
Our notation $(\vec q, q_0)$ refers to the imaginary and real parts of a quaternion respectively.
Thus, a general $[R|T]$ transformation can be represented by 
\begin{align}
Q = T\otimes R = R + \epsilon \frac{1}{2} (\vec a, 0) \otimes R \equiv q + \epsilon q'.
\end{align}

An equation concerning DQ can always be split in two equations, one for the ``primal'' part which is a pure quaternion, and one for the ``dual'' part which is the quaternion coefficient of $\epsilon$.
\section{Hand-eye calibration problem in terms of dual-quaternions}
\label{sec:HandeyeCalibration}
 
The hand-eye calibration problem can be formulated using the following set of equations
 \begin{align}
 \Delta C_i \circ X = X \circ \Delta H_i, \quad i=1,...,N,
 \end{align}
 where $\Delta  C_i$ and $\Delta H_i$ are corresponding relative poses of two rigidly attached sources (see fig.~\ref{fig:hand_eye_setup}), which from now on we shorten as $C_i$ and $H_i$.
 Using the DQ representation the equation can be written as \cite{Daniilidis1999HandEyeCU}
 \begin{align}
 Q_{C_i} \otimes Q_X = Q_X \otimes Q_{H_i}, \quad i=1,...,N,
 \end{align}
 and more explicitly as
 \begin{align}
 &q_{C_i} \otimes q_X = q_X \otimes q_{H_i}, \quad i=1,...,N,\nonumber\\
 &q_{C_i} \otimes q'_X + q'_{C_i} \otimes q_X 
 = q_X \otimes q'_{H_i} + q'_X \otimes q_{H_i},
 \end{align}	 
 where we separated the primal and dual parts to two equations.
 Next we translate the quaternionic multiplications to matrix multiplications using $q\otimes p \equiv \mathbf{L}(q)p \equiv \mathbf{R}(p)q$, where $q$ and $p$ are treated as real four-dimensional vectors on the RHS and $\mathbf{L}, \mathbf{R}$ are $4 \times 4$ matrices, and we get
 \begin{align}
 &\left(\mathbf{L}(q_{C_i}) - \mathbf{R}(q_{H_i})\right)q_X = 0, \quad i=1,...,N,\\
 &\left(\mathbf{L}(q'_{C_i}) - \mathbf{R}(q'_{H_i})\right)q_X + \left(\mathbf{L}(q_{C_i}) - \mathbf{R}(q_{H_i})\right)q'_X = 0\nonumber.
 \end{align}
 For simplicity of notation we define $A_i \equiv \mathbf{L}(q_{C_i}) - \mathbf{R}(q_{H_i})$, $B_i \equiv \mathbf{L}(q'_{C_i}) - \mathbf{R}(q'_{H_i})$ and $q \equiv q_X$, $q' \equiv q'_X$, so we have 
 \begin{align}\label{eq:linear_eq}
 A_i q = 0, \quad &B_i q + A_i q' = 0, \quad i=1,...,N.
 \end{align}
 
 As described in \cite{Daniilidis1999HandEyeCU}, in the noise free case where the equations can be solved exactly, the concatenated $8$-dimensional vector $(q, q')$ is in the null space of $ \tiny \begin{pmatrix} 
 A & 0\\
 B & A
 \end{pmatrix}$, where $A$ and $B$ are a stacks of $A_i$ and $B_i$ respectively.
 Moreover, the relative transformation can only change the screw axis from one frame to another so the angle and screw pitch \dof are omitted in the treatment of \cite{Daniilidis1999HandEyeCU}, which also allows to reduce $A_i$ and $B_i$ to $3\times 4$ matrices instead of $4\times 4$ matrices.
 
 In our analysis we prefer not to cancel the invariant screw parameters since we generally expect noisy data, and these parameters act as sort of relevant weights for our data, thus, we work with the full $4\times 4$ matrices.
 
  \subsection{The minimization problem}\label{sec:minimization_problem}
 
 Our next step is to define the minimization problem. Generally, in the presence of noise, the RHS of (\ref{eq:linear_eq}) will differ from zero.
 We define our minimization problem as the simultaneous minimization of the squared $L_2$ norm of residuals of the two equations in (\ref{eq:linear_eq}), namely
 \begin{align}\label{eq:opt_prob}
 &\arg\min_{q,q'}\sum_{i=1}^N\left(|A_i q|^2 + \alpha^2 |B_i q + A_i q'|^2\right),\quad \nonumber\\
 &\text{subject to: } |q|=1, \quad q\cdot q' = 0,
 \end{align}
 where we introduced a fixed dimensionful parameter $\alpha\in \mathbb{R}$ with $1/\text{length}$ units, which may account for the different units of the two terms. For simplicity we will treat $\alpha$  as a fixed tunable parameter. More generally one could use a covariance matrix inside the norm based on specific measurements without changing our analysis.
 
 Our choice for the cost function is natural given the dual-quaternion formulation, however one can find plenty of other non-equivalent choices in the literature using different metrics on $SE(3)$, see \cite{Esquivel2015EyetoeyeCE} for detailed discussion.

 \subsubsection{Minimization equations}
 In this section we present the minimization problem in more details, and set up the notation for the rest of the paper.
 First, we express the cost function (\ref{eq:opt_prob}) using two Lagrange multipliers corresponding to the two constraints,
 \begin{align}\label{eq:cost_function}
 L\! = &\!\! \sum_i \left(|A_i q|^2 + \alpha^2 |B_i q + A_i q'|^2\right)\! + \lambda(1-q^2)\! -2 \mu q\!\cdot\! q'
\nonumber\\  = & ~q^T S q\! + \! q'^T M q'\! + \! 2 q^T W q'\!
+ \!\lambda(1-q^2)\! - \!2 \mu q\! \cdot\! q',
 \end{align}
 where $\lambda$ and $\mu$ are Lagrange multipliers, and $S \equiv A^T A + \alpha^2 B^T B$, $M = \alpha^2 A^T A$ and $W \equiv \alpha^2 B^T A$.
 The minimization equations yield
 \begin{align}\label{eq:minimization_eq1}
 &\frac{1}{2}\frac{\partial L}{\partial q} = 
 S q+W q'-\lambda q-\mu q'=0,\nonumber\\ 
 &\frac{1}{2}\frac{\partial L}{\partial q'} = 
 M q'+W^T q-\mu q=0,
 \end{align}
 along with the two constraint equations. 
 These equations imply that $L = \lambda$ on the solution.
 
 Throughout the paper we assume that the matrix $M$ is full rank, which is the case of interest where the input data is noisy, otherwise one can solve analytically along the lines of \cite{Daniilidis1999HandEyeCU}.
 Thus, we have
 \begin{align}\label{eq:qp}
 q' = M^{-1}(\mu - W^T)q.
 \end{align}	
 Plugging $q'$ in the first equation yields
 \begin{align}\label{eq:ev1}
 Z(\mu)q = \left(Z_0 + \mu Z_1 - \mu^2 Z_2\right)q =\lambda q,
 \end{align}
 where we introduce the notation
 \begin{align}
 Z_0 \equiv ~& S - W M^{-1}W^T,\quad
 Z_1 \equiv ~ W M^{-1} + M^{-1} W^T,\nonumber\\
 Z_2 \equiv ~& M^{-1}.
 \end{align}
 The first constraint equations $|q|^2 = 1$ can be trivially satisfied by normalizing the solution, while the second constraint $q\cdot q' = 0$ implies that
 \begin{align}\label{eq:mu_eq}
 \mu = \frac{1}{2}\frac{q^T Z_1 q}{q^T Z_2 q}.
 \end{align}
 While we can easily solve (\ref{eq:ev1}) as an eigenvalue problem for a given $\mu$, (\ref{eq:mu_eq}) will generally not be satisfied. 
 Unfortunately plugging (\ref{eq:mu_eq}) in (\ref{eq:ev1}) results in a highly non-linear problem.
 
 \subsubsection{Properties of the minimization equations}\label{sec:prorp_of_problem}
 In order to solve the problem it is useful to notice some properties of (\ref{eq:ev1}) and (\ref{eq:mu_eq}).
 By construction, the matrix $Z(\mu)$ is real and symmetric for any real $\mu$, so its eigenvalues are real.
 We further notice that $Z_2$ is positive semi-definite (since it is the inverse of $A^T A$) and so is $Z_0$\footnote{Notice that $Z_0 = A^T A + \alpha^2 B^T C B$, where $C\equiv (1_{4N\times 4N} - A(A^T A)^{-1}A^T)$ is positive semi-definite having four zero eigenvalues and the rest are ones. The four zero eigenvalues correspond to the columns of $A$ (notice that $(A^T A)^{-1}A^T$ is the pseudo-inverse of $A$), and then since the rank of $A(A^T A)^{-1}A^T$ is 4, the rest of the eigenvalues will be one (corresponding to the vectors in the null space of $C$).}.
 
\begin{figure}[htbp]
\begin{subfigure}[b]{0.47\columnwidth}
\includegraphics[width=\linewidth]{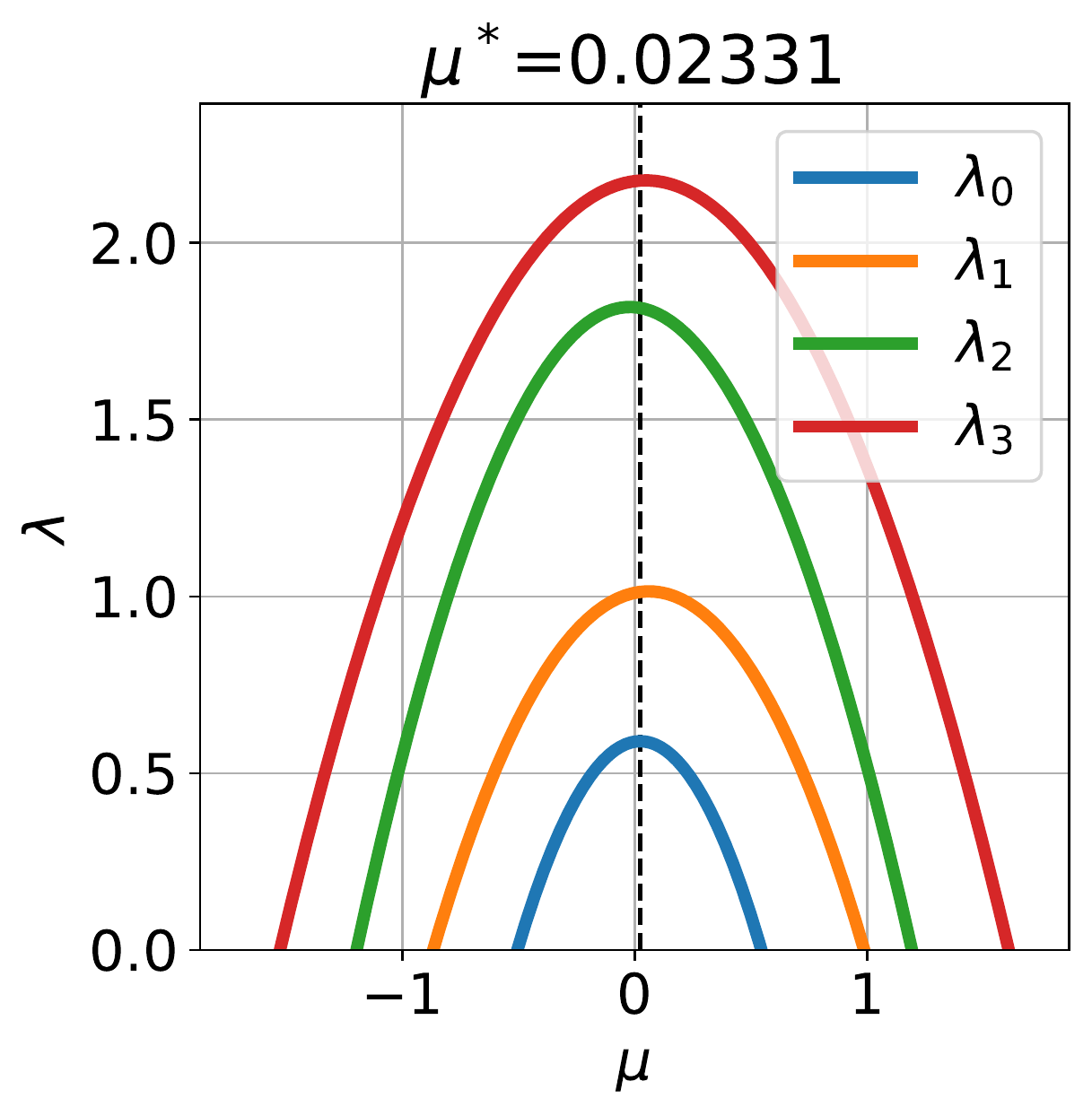} 
\caption{Typical $\lambda(\mu)$}
\label{fig:eigenvalues_normal}
\end{subfigure}
\hfill
\begin{subfigure}[b]{0.47\columnwidth}
\includegraphics[width=\linewidth]{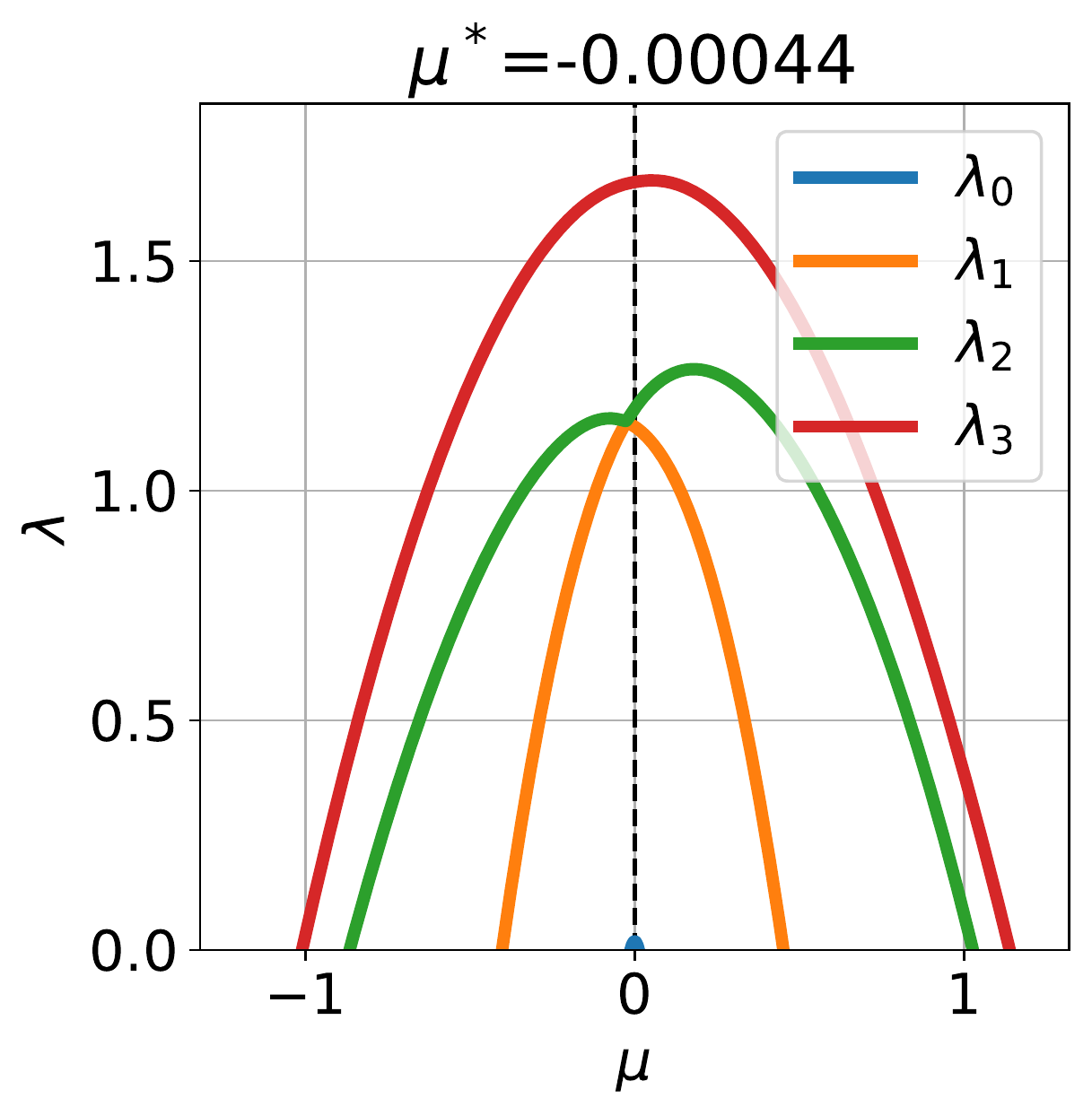}
\caption{Less typical $\lambda(\mu)$}
\label{fig:eigenvalues_multi}
\end{subfigure}
\hfill
\begin{subfigure}[b]{0.47\columnwidth}
\includegraphics[width=\linewidth]{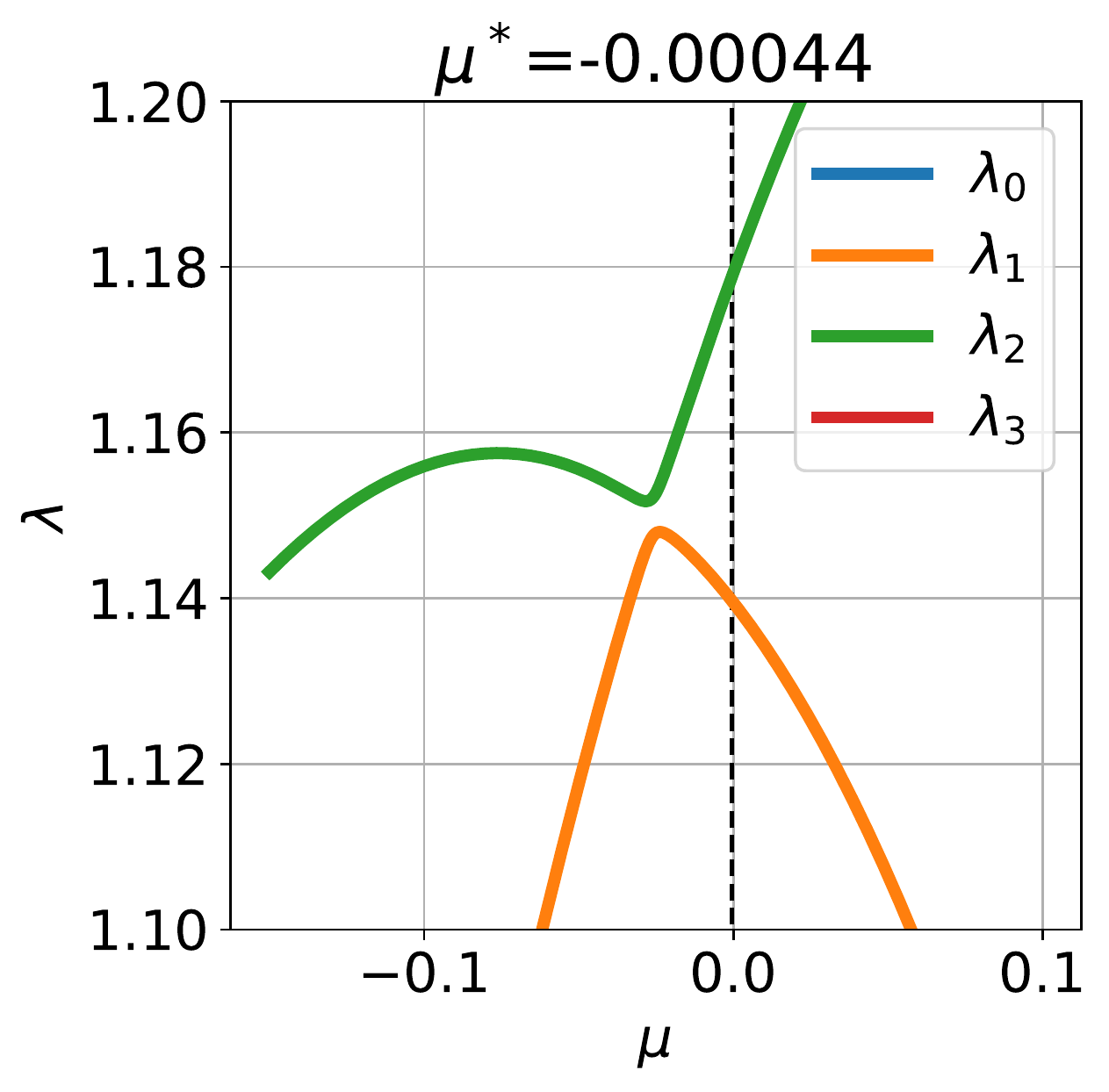}
\caption{Less typical $\lambda(\mu)$ (zoomed)}
\label{fig:zoom}
\end{subfigure}
\hfill
\begin{subfigure}[b]{0.47\columnwidth}
\includegraphics[width=\linewidth]{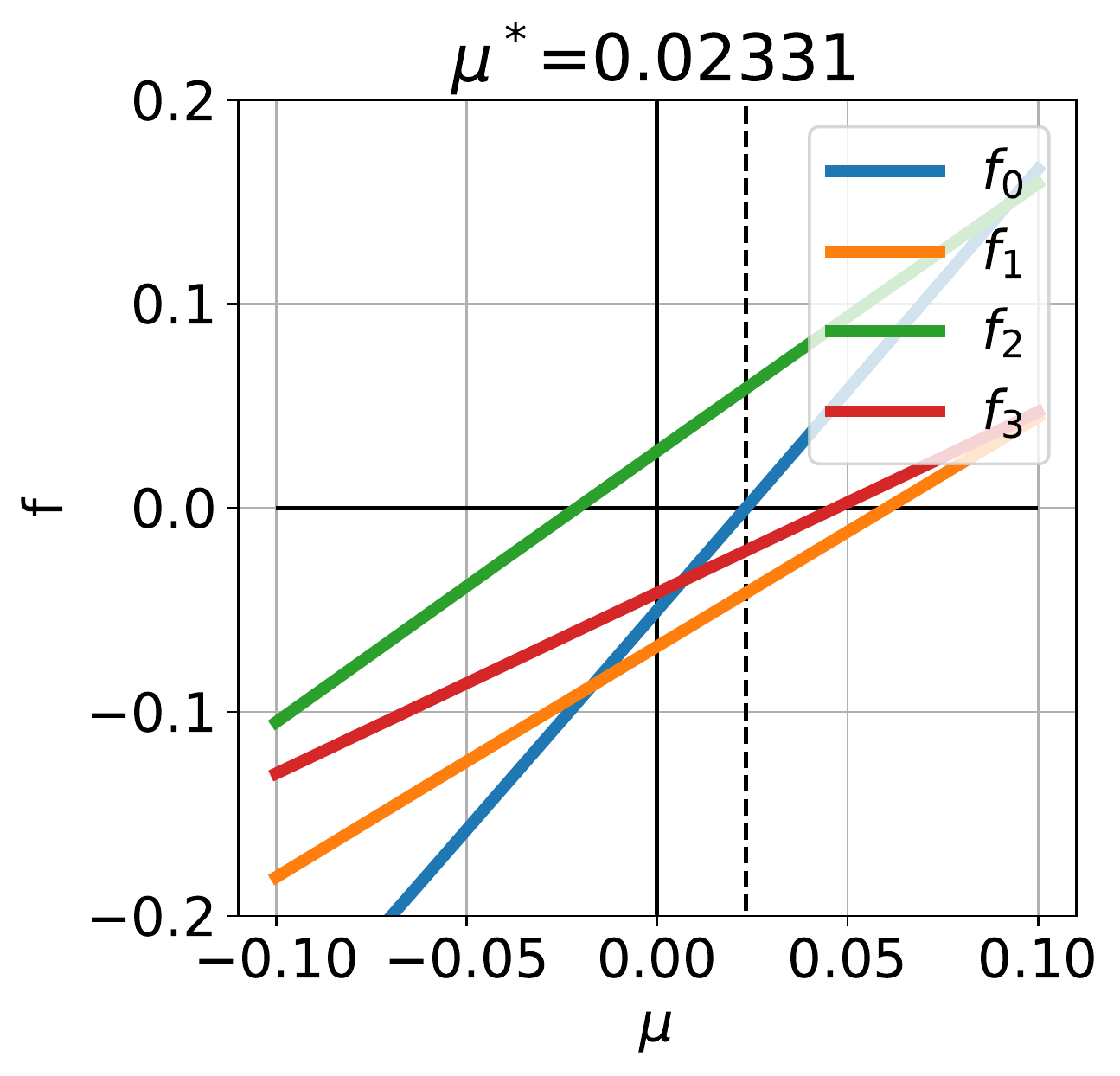}
\caption{Typical $f(\mu)$}
\label{fig:nice_f_of_mu}
\end{subfigure}
\caption{\small Lagrange multipliers space.
Subscripts in $\lambda_i(\mu)$ and $f_i(\mu)$ indicate the $i$-th EV for a given $\mu$ in ascending order.
The optimal $\mu$ is denoted by $\mu^*$ and the dashed black line. In (a) and (b) the optimal solution corresponds to the maximum of the blue line.
In (b) the number of real roots starts at 8, reduce to 6, 4 and then grows to 6 and reduces to 4, 2 and finally to 0. We also see the level repulsion effect where the eigenvalues almost intersect but eventually repel each other, see (c).
  In (d) solution $\mu^*$ corresponds to the intersection of $f_0$ with the $\mu$-axis (same data as in (a)).
  }
\label{fig:LM}
\end{figure}

 Considering these properties, the eigenvalue curves $\lambda_i(\mu)$ (each of the four roots ordered by their value) in (\ref{eq:ev1}) 
 generally should
 not intersect as we change $\mu$, as a consequence of the von Neumann-Wigner theorem\footnote{The probability to get an eigenvalue intersection for generic matrices (due to noise) in our problem when changing $\mu$ is practically zero. We also observed this phenomena in our experiments, see fig.~\ref{fig:zoom}.} \cite{1929PhyZ...30..467V}, \cite{lax07}.
 
 A useful way to think of the problem is in terms of the Lagrange multipliers parameter space, given by a real algebraic curve defined by (\ref{eq:ev1}),
 \begin{align}\label{eq:alg_crv}
   p(\lambda, \mu) = \det (Z(\mu) - \lambda) = \sum_{m=0}^4 \lambda^m c_{8-2m}(\mu) = 0,  
 \end{align}
which is a 4-th degree polynomial in $\lambda$ and 8-th degree in $\mu$, and $c_{n}(\mu)$ is an n-th degree real polynomial in $\mu$ (see figure \ref{fig:LM}). The functions $\lambda_i(\mu)$ can be found analytically, but their explicit expressions are not very illuminating.

The constraint $q(\mu)\cdot q'(\mu) = 0$ dictates which points on the curves $\lambda_i(\mu)$ correspond to the solution. 
 A key observation is that these are the extrema points of $\lambda_i(\mu)$. To show this we use the fact that $\lambda = q^T Z(\mu) q$, so 
 \begin{align}
\frac{\partial\lambda}{\partial\mu}\bigg|_{\text{sol}} & = 2 \dot q^T Z(\mu) q + q^T \dot Z(\mu) q\\
 & = 2 \lambda \dot q^T q - 2 q^T M^{-1}(\mu - W^T) q
   = -2 q^T q'\nonumber,
 \end{align}
where we used (\ref{eq:ev1}) so we are on the solution, and $q^2=1$ which implies $\dot q^T q = 0$, and a dot denotes a $\mu$ derivative.

Thus, recalling that $L=\lambda$ on the solution, (\ref{eq:opt_prob}) can be solved by finding the extrema points of $\lambda(\mu)$ and picking the one with the lowest value. This can be done by finding the real intersection points of the polynomials $p(\lambda, \mu)$ and  $\partial_\mu p(\lambda, \mu)$, which in turn is equivalent to finding the multiplicities of the 8-th degree polynomial in $\mu$, as a function of its $\lambda$ dependent coefficients. In either case, this is equivalent to the vanishing of the determinant of the Sylvester matrix (the resultant) of the two polynomials $p(\lambda, \mu)$ and  $\partial_\mu p(\lambda, \mu)$ \wrt $\lambda$ or $\mu$.
In both cases the solution is given by the smallest real root of a 1d 28-th degree polynomial.

However, considering further properties of the problem allows to find the solution in more efficient ways.
  We notice that for large $|\mu|$ (\ref{eq:ev1}) and (\ref{eq:qp}) become
 \begin{align}
 -\mu^2 Z_2 q \simeq\lambda q, \quad q' \simeq \mu Z_2 q,
 \end{align}
 so $\lambda_i(|\mu|) \simeq -\mu^2 \xi_i$ where $\xi_i \geq 0$ are eigenvalues of $Z_2$.
 Similarly, $Z(\mu=0) = Z_0$ which is PSD, so $\lambda_i(\mu=0) \geq 0$.
 
 Combining these two observations together with the von Neumann-Wigner non-crossing theorem, each of the four eigenvalues $\lambda_i(\mu)$ must cross
 the $\mu$ axis at least twice.
 However, the number of solutions to $Z(\mu)q = 0$ in terms of $\mu$ is at most eight, so we conclude that every eigenvalue $\lambda_i(\mu)$ crosses the $\mu$ axis exactly twice, say at $\mu_1 < 0 < \mu_2$, and then goes asymptotically quadratically to $-\infty$ for $\mu \gg \mu_2$ and $\mu \ll \mu_1$ (there could be cases where the two points unite and we get one solution at $\mu=0$, which corresponds to the noise free case). 

 Next, we change our point of view and consider $\mu(\lambda)$. We already established that the number of real solutions to $\mu(\lambda=0)$ must be eight, which is the maximum number of solutions.
 As we increase $\lambda$, the number of real solutions can either stay constant or drop by an even integer number\footnote{For a given $\lambda$ we can rewrite the eigenvalue problem in terms of $\mu$ as an $8\times 8$ eigenvalue problem, with a non-symmetric matrix, so in general the properties of $\mu$ are less constrained and $\mu$ can become complex.}.
 The first jump from eight to six solutions corresponds to the first maximum of $\lambda(\mu)$, which is the minimal cost solution to our problem. This must correspond to the maximum of $\lambda_{i=0}(\mu)$, since otherwise the number of real solutions to the $\mu$ polynomial will exceed the maximum number (eight). Multiple extrema are allowed only for higher $\lambda_{i>0}$ functions, see \ref{fig:eigenvalues_multi}.
 Similarly we can look for root multiplicities in terms of the discriminant, since the extrema of $\lambda(\mu)$ correspond to such points.
 
 We conclude that the smallest eigenvalue extrema must corresponds to the unique maxima of $\lambda_0(\mu)$. Note furthermore that $\lambda_0(\mu)$ can be written as the minimum over a set of concave functions\footnote{Let $f_q(\mu) = q^TZ(\mu)q$. Then $\lambda_0(\mu) = \min_{|q| = 1} f_q(\mu)$ and $\partial_\mu^2f_q(\mu)  = -q^TZ_2q \leq 0$ for a fixed $q$ since $Z_2$ is PSD.} and is thus itself concave (or equivalently, $-\lambda_0(\mu)$ is convex). This means that we can efficiently solve the hand-eye calibration problem by finding the maximum of this function and extracting the corresponding $q, q'$.

  \subsection{Solving the minimization problem}

Based on the discussion above, we introduce various strategies to solve the minimization equations, presenting both exact and approximate solution.

 \subsubsection{Optimal 1D line search (DQOpt)}\label{subsec:1dlinesearch}
  This is the simplest algorithm we present which yields the optimal solution. We first define the function
  \begin{align}\label{eq:f_eq}
    f_0(\mu) \equiv q_0(\mu)\cdot q'_0(\mu) = -\frac{1}{2}\frac{d \lambda_0(\mu)}{d \mu},   
  \end{align}
  where $q_0$ is the smallest eigenvalue of $Z(\mu)$, and $q'_0$ is computed using (\ref{eq:qp}).
  As explained above, this is a monotonic function with a unique root, see figure \ref{fig:nice_f_of_mu}.
  To find the optimal solution we first find the root of this function, $f_0(\mu_\ast) = 0$, and solve for $q$ as the eigenvector corresponding to the smallest eigenvalue of $Z(\mu_\ast)$ and for $q'$ using (\ref{eq:qp}).
  Alternatively, one can look for the maximum of $\lambda_0(\mu)$ directly which correspond to the same solutions, see \ref{fig:eigenvalues_normal}.
  
  We further derived alternative algorithms to arrive at the optimal solution, again, using only a 1D line search based on resultants and roots counting. These algorithms yield the same optimal solution, but are slower to evaluate.
  Since they are conceptually very different and provide a different view on the problem we believe it is interesting to include them in this work. However, since we shall only use the above optimal solution throughout the paper we relegate their description to appendix \ref{sec:polystuff}.
  
  \subsubsection{Two steps (DQ2steps)}\label{subsec:2steps}
  The two steps approximation, solving first the rotation equation, takes a very simple form in our formulation.
One first solves for $q$ for finding the eigenvector corresponding to the smallest eigenvalue in
\begin{align}
M q = \lambda_0 q,
\end{align}
and the corresponding $q'$ is given by
\begin{align}\label{eq:qp_sol}
q' = M^{-1}\left(\frac{1}{2}\frac{q^T Z_1 q}{q^T Z_2 q} - W^T\right)q.
\end{align}

 \subsubsection{Convex relaxation (DQCovRlx)}\label{subsec:ConvRlx}
Here we approximate the solution while relaxing the $q\cdot q'=0$ constraint, and then project the solution to the non-convex constraint space using (\ref{eq:mu_eq}), thus we solve 
\begin{align}\label{eq:conv_relax}
Z_0 q = \lambda_0 q,
\end{align}
and the corresponding $q'$ is given by (\ref{eq:qp_sol}) again.
We can also get a nice expression bounding the gap,
\begin{align}
\Delta \lambda = \lambda^c - \lambda_0 = 
\frac{1}{4}\frac{(q^T Z_1 q)^2}{q^T Z_2 q},
\end{align}
where $\lambda^c$ is the constrained solution.
Namely, the true cost for the non-convex problem, $\lambda^\ast$, satisfies $\lambda_0 \leq \lambda^\ast \leq \lambda^c$.

  \subsubsection{Second order approximation (DQ2ndOrd)}\label{subsec:2ndOrd}
  Starting with the relaxed solution (\ref{eq:conv_relax}), corresponding to $\mu=0$ (which is true in the noise free case), we can have an analytic expansion in small $\mu$. A detailed derivation of a recursive formula for the dual-quaternion solution to any order in $\mu$ is given in appendix \ref{app:second_order}. 
  The resulting second order approximation for yields 
  \begin{align}\label{eq:sec_ord}
  \mu^\ast_{(2)} =& \frac{1}{2}\frac{Z_1^{00}}{Z_2^{00} - \sum_a\frac{(Z_1^{a0})^2}{\lambda_{0a}}},\nonumber\\
  q_{(2)} =~ & q_0 + \mu^\ast_{(2)}\sum_a\frac{Z_1^{a0}}{\lambda_{0a}}q_a
  + \mu^{\ast 2}_{(2)}\bigg(-\frac{1}{2}q_0\sum_a\left(\frac{Z_1^{a0}}{\lambda_{0a}}\right)^2\nonumber\\
  & +\sum_a \frac{\sum_b\frac{Z_1^{b0}Z_1^{ab}}{\lambda_{0b}} - Z_2^{a0}-\frac{Z_1^{00}Z_1^{a0}}{\lambda_{0a}}}{\lambda_{0a}}q_a
  \bigg),
  \end{align}
  where the $q$'s and $\lambda$'s are defined by $Z_0 q_a = \lambda_a q_a$. We also introduced the short-hand notation $Z_i^{ab}\equiv q^T_a Z_i q_b$ and $\lambda_{0a}\equiv \lambda_0-\lambda_a$ and all the sums are running over $a=1,2,3$.
  After normalizing $q_{(2)}$ we solve for $q'$ using (\ref{eq:qp_sol}).
  
 Since $\mu$ is not a dimensionless parameter, one might prefer to expand in $\lambda$ which is dimensionless.
 We can start with $\lambda_0=0$ which is the solution in the noise free case.
 However, this leads to an 8-th degree polynomial in $\mu$ when coming to solve (\ref{eq:ev1}) which complicates a bit the procedure.
 Instead, we can expand in $\lambda$ around the relaxed ($\mu=0$) solution, so $\lambda = \lambda_0 + \Delta \lambda$ and the expansion is in $\Delta \lambda$, and $\lambda_0$ corresponds to the smallest eigenvalue of $Z_0$. 
 The second order expansions allowing to find $q$ is slightly more elaborated compared to the previous one and can be found in appendix \ref{app:second_order}. Then as in the previous approximation we solve for $q'$ using (\ref{eq:qp_sol}).

 The results of this expansion are comparable to the previous one in our experiments, though they are not equivalent (both giving very accurate results).
 
  \subsubsection{Iterative solution (DQItr)}\label{subsec:itr}
  Defining the function
  \begin{align}
 \tilde \mu(\mu) = \frac{1}{2}\frac{q(\mu)^T Z_1 q(\mu)}{q(\mu)^T Z_2 q(\mu)},
 \end{align}
 where $q(\mu)$ is the eigenvector corresponding to the  smallest eigenvalue of (\ref{eq:ev1}),
 the solution to the problem is the fixed point, namely $\tilde \mu(\mu) = \mu$.
 Since $\tilde \mu(\mu)$ is bounded as explained in the previous section, iteratively estimating $\tilde \mu$ and plugging the value back convergence to the solution.
 In practice this procedure converges very quickly, though it is unstable in the noise free case.

 \subsection{Adding a prior}
 The hand-eye calibration problem is not always well posed, for example in the case of planar motion some \dof are not fixed, and might require a regularization term for a unique solution. Furthermore, constructing the problem from a statistical model perspective, as a MAP probability estimation requires a prior.
 Thus, it is useful to incorporate a prior term in our formulation.
  
  Let us assume we have some prior knowledge regarding the hand-eye calibration, namely that the unit dual-quaternion representing the calibration is close to some given $\hat Q$.
  We define $Q = \hat Q \otimes \delta Q$, such that $\delta Q = \delta q + \epsilon \delta q'$ is the deviation from the prior, which is expected to be small, namely
  $\delta Q \simeq 1$, or equivalently $\delta q \simeq 1$ and $\delta q' \simeq 0$.
  Thus, we introduce the prior by penalizing large $\delta Q$
  \begin{align}\label{eq:prior1}
  L_{\text{prior}}(q,q') =  L(q,q') + a (1-\delta q)^2 + b \delta q'^2,
  \end{align}
  where $a, b > 0$ are constant weight parameters (notice that $q$ and $\delta q$ are not independent). The rotation term has a linear dependence on $\delta q$. This term complicates our analysis, where the eigenvalue problem needs to be promoted to an $8 \times 8$ instead of the $4\times 4$ problem we had before, and our matrices lose some of their nice properties. Nonetheless, this approach is valid and yields good results.
  
However, we can avoid these issues by noting that $|\delta q| = 1$,
$\delta q \simeq 1$ is equivalent to the imaginary part being small. So we instead modify (\ref{eq:prior1}) to
 \begin{align}\label{eq:prior2}
 L_{\text{prior}}\! =\! L + a \delta q^T G \delta q + b \delta q'^2,\!\! \quad G\! =\! \text{diag}(1,\! 1,\! 1,\! 0).
 \end{align}
 We can minimize this cost in the same way we minimize \eqref{eq:cost_function} by noting that
 \begin{align}
  \delta Q = \hat Q^\ast \otimes Q = \mathbf{L}(\hat q^\ast) q + \epsilon(\mathbf{L}(\hat  q'^\ast) q + \mathbf{L}(\hat q^\ast) q').
 \end{align}
 Thus, the two new terms are given by
 \begin{align}
 \delta q^T G \delta q& = q^T \mathbf{L}^{\ast T} G  \mathbf{L}^\ast q, \nonumber\\
 \delta q'^2& = q^T \mathbf{L}'^{\ast T} \mathbf{L}'^\ast q + 2 q^T \mathbf{L}'^{\ast T} \mathbf{L}^\ast q' + q'^T q'\nonumber\\
& = |\hat q'|^2 q^T q + 2 q^T \tilde W q' + q'^T q',
 \end{align}
 where $\tilde W = - \tilde W^T = \mathbf{L}'^{\ast T} \mathbf{L}^\ast = \mathbf{L}(\hat t)$. Because of our constraints, the term $|\hat q'|^2 q^T q$ is a constant which does not change the cost function, so we can ignore it (notice that by doing that the cost function is not guaranteed to be positive anymore, but is bounded from below by $-b |\hat q'|^2$).
  Thus, we can solve \eqref{eq:prior2} in the same way as (\ref{eq:cost_function}) by identifying
 \begin{align}
  S \to ~& S + a \mathbf{L}^{\ast T} G \mathbf{L}^\ast, \nonumber\\
  W \to ~& W + b \tilde W, \nonumber\\
  M \to ~& M+b \mathbf{1}_{4\times 4}.
 \end{align}
 Notice that these modifications to (\ref{eq:cost_function}) do not change the matrix properties discussed above regarding semi-positive definiteness and symmetries.
\section{Experiments}\label{sec:Experiments}
In this work we perform two kinds of experiments. First we compare our optimal algorithm results with a nonlinear optimization implementation to show that our algorithm results in the optimal solution.
Afterwards we compare the performance of our algorithms with other algorithms, on real and synthetic datasets.

Throughout this section we use the algorithms abbreviations introduced in sec.~\ref{sec:HandeyeCalibration}, with the addition of \textbf{Dan} \cite{Daniilidis1999HandEyeCU}, as well as \textbf{ChVecOpt} and \textbf{QuatVecOpt} which minimize 
\begin{align}\label{eq:other_costs}
&\textstyle\sum_i\left(|R^i_C R\! -\! R R^i_H|^2 \!+\! \alpha^2 |(R^i_C\! -\! 1)T\! -\! R T^i_H\! +\! T^i_C|^2\right),\nonumber\\
&\textstyle\sum_i\left(|q^i_C q\! -\! q q^i_H|^2 \!+\! \alpha^2 |(R^i_C\! -\! 1)T\! -\! R T^i_H\! +\! T^i_C|^2\right)
\end{align}
\wrt $R,T$ and $q,T$ using nonlinear optimization respectively.
Finally, \textbf{DQNLOpt} and \textbf{DQNLOptRnd} represent nonlinear optimization of our cost function (\ref{eq:cost_function}) initialized with \textbf{DQOpt} and randomly respectively.

All the algorithms are implemented using python with numpy and scipy optimize packages for 1D root finding.

\begin{figure*}[htpb]
\centering
\includegraphics[width=0.98\linewidth]{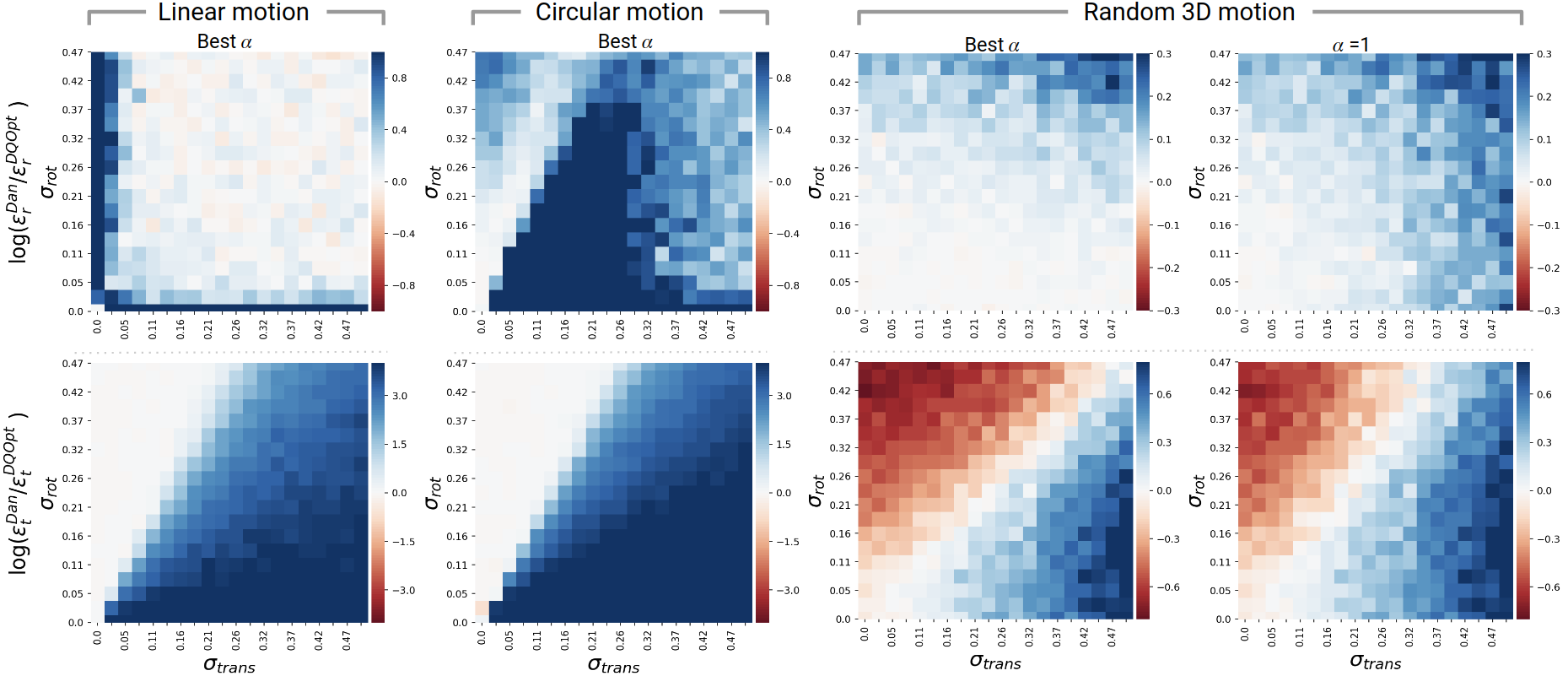} 
\caption{\small \textbf{Synthetic experiments}. We compare between \textbf{Dan} and \textbf{DQOpt} on the three different motions by showing the error response of the solvers when the relative poses are subject to increasing rotational and translational noise (sec.~\ref{subsec:exp_setup_syn}). The heat maps show the mean ratio (60 iterations) of the error responses (first row - rotation, second row - translation). A blue (red) color indicates that \textbf{DQOpt} has $e^{\text{value}}$ times lower (higher) error compared to \textbf{Dan}.
It is possible to notice how \textbf{DQOpt} dramatically outperform \textbf{Dan} on degenerate motions (linear and circular). For the 3D case (column 3) both \textbf{DQOpt} and \textbf{Dan} show high accuracy (see table \ref{tb:real_data_exp}) and the error ratio loses its significance (\ie they both perform well). In the last column, we compare the two solvers for a $\alpha=1$. }
\label{fig:experiments_synt}
\end{figure*}

\subsection{Synthetic experiments}\label{subsec:exp_setup_syn}
Our synthetic dataset consists of three scenarios:
 \\ - \textbf{Random} - uniformly sampled from $SO(3) \times E(3)$. \\ - \textbf{Line} - straight motion + small 3d perturbations.
 \\ - \textbf{Circle} - 2d circular motion + small 3d perturbations. \\
The relative poses are related by the extrinsic transformation with addition of noise.
For the translations we use multivariate Gaussian noise, while for the rotations we use a uniform distribution on $S^2$ for the axis and a Gaussian distribution for the angle (more details can be found in appendix \ref{app:datasets}).

\subsection{Real data experiments}\label{subsec:exp_setup_real}
We recorded poses from two rigidly attached ZED cameras running the built-in visual odometry. The cameras were moved by hand, creating a general 3d motion scenario and three approximately planar scenarios: general, straight, and circular. The collected data is significantly noisy due to the inaccuracy of the tracking algorithm and time synchronization issues. More details on the data acquisition process and pre-filtering can be found in appendix \ref{app:datasets}.

\subsection{Weighting factor} \label{subsec:weighting_factor}
The proper choice of the weighting factor $\alpha$, which was introduced in sec.~\ref{sec:minimization_problem},
depends on the noise model.
In order to have a fair comparison we add the $\alpha$ parameter to all the algorithm implementations, see (\ref{eq:other_costs}), and by scaling the translation part by $\alpha$ for \textbf{Dan}.
We introduce $\alpha_{\text{best}}$ for rotation and translation, which are the values that give the best results in terms of the rotation and translation mean error respectively \wrt the ground-truth. Note that these values are not necessarily the same for all algorithms and even for the same dataset.
In this way we treat all the algorithms in a fair way in contrast to setting one value which might benefit certain algorithms more than others.

\begin{table}[h!]
\scriptsize
\setlength\tabcolsep{2.5pt}
 \begin{tabular}{||l | c | c | c | c||} 
 \hline
 \rowcolor{ColorTable}
 \multicolumn{5}{||c||}{Relative cost difference $\frac{alg - DQOpt}{alg + DQOpt}$} \\
 \hline
 \rowcolor{ColorTable}
 alg & mean & std & min & max\\ [0.5ex] 
 \hline
 DQNLOpt & $6.3\times 10^{-18}$ & $5.8\times 10^{-16}$ & $-3.0\times 10^{-15}$ & $2.8\times 10^{-15}$ \\ 
 \hline
 DQNLOptRnd & $6.8\times 10^{-2}$ & $1\times 2.1^{-1}$ & $4.3\times 10^{-12}$ & $8.9\times 10^{-1}$ \\ 
 \hline
 DQ2ndOrd & $6.6\times 10^{-10}$ & $5.4\times 10^{-9}$ & $-2.3\times 10^{-15}$ & $1.3\times 10^{-7}$ \\
 \hline
 DQConvRlx & $2.7\times 10^{-5}$ & $7.0\times 10^{-5}$ & $2.15\times 10^{-12}$ & $5.3\times 10^{-4}$ \\
 \hline
 DQ2Steps & $8.4\times 10^{-3}$ & $1.6\times 10^{-2}$ & $2.4\times 10^{-5}$ & $9.7\times 10^{-2}$ \\
 \hline
 \hline
 \rowcolor{ColorTable}
 \multicolumn{5}{||c||}{Timing in $\mu$sec} \\
 \hline
 \rowcolor{ColorTable}
 alg & mean  & std & min & max\\ [0.5ex] 
 \hline
 DQOpt & 299&   92&  158& 1199 \\ 
 \hline
 DQ2ndOrd & 155&  53&  98& 818 \\
 \hline
 Dan & 92&   33&   61& 1200\\ 
 \hline
 DQConvRlx & 77&  36&  43& 896\\
 \hline
 DQ2Steps & 63&   23&   42& 1242\\
 \hline
 \end{tabular}
 \caption{\small Upper table: Results for the (signed) relative difference (averaged over 2450 real data runs) between \textbf{DQOpt} and our other algorithms. Positive values imply that the competing algorithm has a higher cost function.
 Lower table: Algorithm's timing averaged over 24500 real data runs.}
\label{tb:optimality}
\end{table}

\newcolumntype{s}{>{\columncolor[HTML]{AAACED}}}
\arrayrulecolor[HTML]{999999}

\begin{table*}[h!]
\begin{subtable}[t]{1\textwidth}
\scriptsize
\setlength\tabcolsep{6pt}
\begin{center}
\def\arraystretch{1.5}
\begin{tabular}{|| ll | lll | lll | lll | lll ||}
\hline 
\rowcolor{ColorTable}
    &
    & \multicolumn{3}{c|}{3d motion}  
    & \multicolumn{3}{c|}{Rotation} 
    & \multicolumn{3}{c|}{Planar} 
    & \multicolumn{3}{c||}{Linear}
\\
\rowcolor{ColorTable}
 $\alpha$ & Algorithm 
 & $\varepsilon_{\text{r}}$ (deg) & $\varepsilon_{\text{t}}$ (cm) & $\alpha$
 & $\varepsilon_{\text{r}}$ (deg) & $\varepsilon_{\text{t}}$ (cm) & $\alpha$
 & $\varepsilon_{\text{r}}$ (deg) & $\varepsilon_{\text{t}}$ (cm) & $\alpha$
 & $\varepsilon_{\text{r}}$ (deg) & $\varepsilon_{\text{t}}$ (cm) & $\alpha$
\\
\hline
\multirow{7}{0pt}{\rotatebox{90}{best rotation $\alpha~~~~~~~$}}
& DQOpt
&2.03$\pm$ $\!\! \tiny\text{0.73}\atop \!\! \text{0.60}$&5.32$\pm$ $\!\! \tiny\text{1.95}\atop \!\! \text{1.39}$&1.47
&\textbf{6.91}$\pm$ $\!\! \tiny\text{\textbf{5.45}}\atop \!\! \text{\textbf{3.47}}$&\textbf{5.05}$\pm$ $\!\! \tiny\text{\textbf{3.08}}\atop \!\! \text{\textbf{1.56}}$&0.20
&2.55$\pm$ $\!\! \tiny\text{0.44}\atop \!\! \text{0.41}$&17.7$\pm$ $\!\! \tiny\text{31.4}\atop \!\! \text{9.61}$&0.37
&\textbf{2.89}$\pm$ $\!\! \tiny\text{\textbf{0.22}}\atop \!\! \text{\textbf{0.17}}$&26.3$\pm$ $\!\! \tiny\text{10.5}\atop \!\! \text{9.03}$&1.14
\\& DQ2ndOrd
&2.03$\pm$ $\!\! \tiny\text{0.73}\atop \!\! \text{0.60}$&5.32$\pm$ $\!\! \tiny\text{1.95}\atop \!\! \text{1.39}$&1.47
&\textbf{6.91}$\pm$ $\!\! \tiny\text{\textbf{5.45}}\atop \!\! \text{\textbf{3.47}}$&\textbf{5.05}$\pm$ $\!\! \tiny\text{\textbf{3.08}}\atop \!\! \text{\textbf{1.56}}$&0.20
&2.55$\pm$ $\!\! \tiny\text{0.44}\atop \!\! \text{0.41}$&17.7$\pm$ $\!\! \tiny\text{31.4}\atop \!\! \text{9.61}$&0.37
&\textbf{2.89}$\pm$ $\!\! \tiny\text{\textbf{0.22}}\atop \!\! \text{\textbf{0.17}}$&26.3$\pm$ $\!\! \tiny\text{10.5}\atop \!\! \text{9.03}$&1.14
\\& DQConvRlx
&\textbf{2.00}$\pm$ $\!\! \tiny\text{\textbf{0.76}}\atop \!\! \text{\textbf{0.58}}$&5.32$\pm$ $\!\! \tiny\text{1.95}\atop \!\! \text{1.40}$&1.47
&6.95$\pm$ $\!\! \tiny\text{5.39}\atop \!\! \text{3.50}$&\textbf{5.05}$\pm$ $\!\! \tiny\text{\textbf{3.10}}\atop \!\! \text{\textbf{1.56}}$&0.20
&\textbf{2.53}$\pm$ $\!\! \tiny\text{\textbf{0.77}}\atop \!\! \text{\textbf{0.51}}$&\textbf{17.2}$\pm$ $\!\! \tiny\text{\textbf{27.3}}\atop \!\! \text{\textbf{9.64}}$&0.17
&\textbf{2.89}$\pm$ $\!\! \tiny\text{\textbf{0.23}}\atop \!\! \text{\textbf{0.18}}$&\textbf{25.9}$\pm$ $\!\! \tiny\text{\textbf{10.8}}\atop \!\! \text{\textbf{9.01}}$&1.04
\\& DQ2Steps
&3.26$\pm$ $\!\! \tiny\text{1.24}\atop \!\! \text{0.99}$&\textbf{5.27}$\pm$ $\!\! \tiny\text{\textbf{2.05}}\atop \!\! \text{\textbf{1.52}}$& - 
&17.0$\pm$ $\!\! \tiny\text{13.2}\atop \!\! \text{9.09}$&7.59$\pm$ $\!\! \tiny\text{4.12}\atop \!\! \text{2.86}$& - 
&7.17$\pm$ $\!\! \tiny\text{6.77}\atop \!\! \text{3.75}$&19.5$\pm$ $\!\! \tiny\text{23.1}\atop \!\! \text{10.1}$& - 
&9.46$\pm$ $\!\! \tiny\text{8.92}\atop \!\! \text{3.20}$&43.9$\pm$ $\!\! \tiny\text{45.8}\atop \!\! \text{17.4}$& - 
 
\\& Dan \cite{Daniilidis1999HandEyeCU}
&\textbf{2.00}$\pm$ $\!\! \tiny\text{\textbf{0.75}}\atop \!\! \text{\textbf{0.60}}$&5.45$\pm$ $\!\! \tiny\text{2.14}\atop \!\! \text{1.54}$&1.47
&11.4$\pm$ $\!\! \tiny\text{9.91}\atop \!\! \text{4.99}$&27.4$\pm$ $\!\! \tiny\text{73.5}\atop \!\! \text{20.4}$&17.8
&4.01$\pm$ $\!\! \tiny\text{4.20}\atop \!\! \text{1.20}$&565.$\pm$ $\!\! \tiny\text{1132}\atop \!\! \text{336.}$&1.90
&18.3$\pm$ $\!\! \tiny\text{26.5}\atop \!\! \text{10.7}$&5801$\pm$ $\!\! \tiny\text{4347}\atop \!\! \text{2592}$&2.26
\\& ChVecOpt
&2.08$\pm$ $\!\! \tiny\text{0.74}\atop \!\! \text{0.63}$&5.96$\pm$ $\!\! \tiny\text{1.79}\atop \!\! \text{1.55}$&2.47
&7.22$\pm$ $\!\! \tiny\text{5.00}\atop \!\! \text{3.72}$&28.0$\pm$ $\!\! \tiny\text{18.0}\atop \!\! \text{14.8}$&0.31
&2.75$\pm$ $\!\! \tiny\text{0.44}\atop \!\! \text{0.44}$&45.3$\pm$ $\!\! \tiny\text{22.4}\atop \!\! \text{22.1}$&4.13
&2.95$\pm$ $\!\! \tiny\text{0.25}\atop \!\! \text{0.19}$&39.6$\pm$ $\!\! \tiny\text{23.6}\atop \!\! \text{15.9}$&1.47
\\& QuatVecOpt
&2.04$\pm$ $\!\! \tiny\text{0.81}\atop \!\! \text{0.56}$&5.71$\pm$ $\!\! \tiny\text{2.06}\atop \!\! \text{1.47}$&0.81
&7.37$\pm$ $\!\! \tiny\text{5.05}\atop \!\! \text{4.06}$&16.9$\pm$ $\!\! \tiny\text{11.0}\atop \!\! \text{8.16}$&0.12
&2.69$\pm$ $\!\! \tiny\text{0.41}\atop \!\! \text{0.37}$&45.8$\pm$ $\!\! \tiny\text{23.0}\atop \!\! \text{24.4}$&1.00
&2.95$\pm$ $\!\! \tiny\text{0.27}\atop \!\! \text{0.18}$&38.8$\pm$ $\!\! \tiny\text{25.8}\atop \!\! \text{16.1}$&0.48
\\\hline\multirow{7}{0pt}{\rotatebox{90}{best translation $\alpha~~~~~~~$}}
& DQOpt
&3.21$\pm$ $\!\! \tiny\text{1.16}\atop \!\! \text{0.99}$&5.26$\pm$ $\!\! \tiny\text{1.99}\atop \!\! \text{1.46}$&0.03
&7.75$\pm$ $\!\! \tiny\text{5.27}\atop \!\! \text{3.98}$&\textbf{5.31}$\pm$ $\!\! \tiny\text{\textbf{2.77}}\atop \!\! \text{\textbf{1.80}}$&0.22
&3.84$\pm$ $\!\! \tiny\text{2.92}\atop \!\! \text{1.89}$&18.5$\pm$ $\!\! \tiny\text{21.1}\atop \!\! \text{10.3}$&0.05
&3.37$\pm$ $\!\! \tiny\text{0.35}\atop \!\! \text{0.29}$&23.4$\pm$ $\!\! \tiny\text{10.4}\atop \!\! \text{7.77}$&17.8
\\& DQ2ndOrd
&3.21$\pm$ $\!\! \tiny\text{1.16}\atop \!\! \text{0.99}$&5.26$\pm$ $\!\! \tiny\text{1.99}\atop \!\! \text{1.46}$&0.03
&7.75$\pm$ $\!\! \tiny\text{5.27}\atop \!\! \text{3.98}$&\textbf{5.31}$\pm$ $\!\! \tiny\text{\textbf{2.77}}\atop \!\! \text{\textbf{1.80}}$&0.22
&3.84$\pm$ $\!\! \tiny\text{2.92}\atop \!\! \text{1.89}$&18.5$\pm$ $\!\! \tiny\text{21.1}\atop \!\! \text{10.3}$&0.05
&3.37$\pm$ $\!\! \tiny\text{0.35}\atop \!\! \text{0.29}$&23.4$\pm$ $\!\! \tiny\text{10.4}\atop \!\! \text{7.77}$&17.8
\\& DQConvRlx
&3.21$\pm$ $\!\! \tiny\text{1.16}\atop \!\! \text{0.99}$&5.26$\pm$ $\!\! \tiny\text{1.99}\atop \!\! \text{1.46}$&0.03
&\textbf{7.72}$\pm$ $\!\! \tiny\text{\textbf{5.38}}\atop \!\! \text{\textbf{3.94}}$&5.34$\pm$ $\!\! \tiny\text{2.74}\atop \!\! \text{1.82}$&0.22
&\textbf{3.83}$\pm$ $\!\! \tiny\text{\textbf{2.89}}\atop \!\! \text{\textbf{1.89}}$&18.4$\pm$ $\!\! \tiny\text{21.1}\atop \!\! \text{10.3}$&0.05
&\textbf{3.35}$\pm$ $\!\! \tiny\text{\textbf{0.36}}\atop \!\! \text{\textbf{0.28}}$&\textbf{23.3}$\pm$ $\!\! \tiny\text{\textbf{10.3}}\atop \!\! \text{\textbf{7.75}}$&17.8
\\& DQ2Steps
&3.26$\pm$ $\!\! \tiny\text{1.24}\atop \!\! \text{0.99}$&5.27$\pm$ $\!\! \tiny\text{2.05}\atop \!\! \text{1.52}$& - 
&17.0$\pm$ $\!\! \tiny\text{13.2}\atop \!\! \text{9.09}$&7.59$\pm$ $\!\! \tiny\text{4.12}\atop \!\! \text{2.86}$& - 
&7.17$\pm$ $\!\! \tiny\text{6.77}\atop \!\! \text{3.75}$&19.5$\pm$ $\!\! \tiny\text{23.1}\atop \!\! \text{10.1}$& - 
&9.46$\pm$ $\!\! \tiny\text{8.92}\atop \!\! \text{3.20}$&43.9$\pm$ $\!\! \tiny\text{45.8}\atop \!\! \text{17.4}$& - 

\\& Dan \cite{Daniilidis1999HandEyeCU}
&\textbf{2.36}$\pm$ $\!\! \tiny\text{\textbf{0.94}}\atop \!\! \text{\textbf{0.72}}$&\textbf{5.14}$\pm$ $\!\! \tiny\text{\textbf{2.07}}\atop \!\! \text{\textbf{1.43}}$&3.19
&10.7$\pm$ $\!\! \tiny\text{11.2}\atop \!\! \text{4.72}$&20.2$\pm$ $\!\! \tiny\text{55.2}\atop \!\! \text{13.9}$&32.5
&4.19$\pm$ $\!\! \tiny\text{5.62}\atop \!\! \text{1.18}$&270.$\pm$ $\!\! \tiny\text{445.}\atop \!\! \text{141.}$&42.1
&19.3$\pm$ $\!\! \tiny\text{27.8}\atop \!\! \text{11.5}$&5208$\pm$ $\!\! \tiny\text{3804}\atop \!\! \text{2110}$&25.1
\\& ChVecOpt
&3.20$\pm$ $\!\! \tiny\text{1.14}\atop \!\! \text{0.98}$&5.88$\pm$ $\!\! \tiny\text{1.84}\atop \!\! \text{1.67}$&0.03
&15.0$\pm$ $\!\! \tiny\text{12.9}\atop \!\! \text{8.79}$&15.9$\pm$ $\!\! \tiny\text{8.88}\atop \!\! \text{6.63}$&0.03
&7.21$\pm$ $\!\! \tiny\text{6.77}\atop \!\! \text{3.71}$&18.5$\pm$ $\!\! \tiny\text{15.2}\atop \!\! \text{8.90}$&0.01
&5.74$\pm$ $\!\! \tiny\text{1.49}\atop \!\! \text{1.59}$&24.6$\pm$ $\!\! \tiny\text{9.79}\atop \!\! \text{7.43}$&0.07
\\& QuatVecOpt
&3.17$\pm$ $\!\! \tiny\text{1.14}\atop \!\! \text{1.09}$&5.69$\pm$ $\!\! \tiny\text{1.93}\atop \!\! \text{1.65}$&0.13
&9.82$\pm$ $\!\! \tiny\text{8.03}\atop \!\! \text{5.64}$&14.8$\pm$ $\!\! \tiny\text{9.11}\atop \!\! \text{6.80}$&0.06
&6.45$\pm$ $\!\! \tiny\text{5.12}\atop \!\! \text{3.29}$&19.9$\pm$ $\!\! \tiny\text{13.5}\atop \!\! \text{8.62}$&0.01
&4.94$\pm$ $\!\! \tiny\text{1.47}\atop \!\! \text{1.27}$&24.2$\pm$ $\!\! \tiny\text{8.55}\atop \!\! \text{7.08}$&0.07
\\\hline
\end{tabular}
\end{center}

\end{subtable}

\bigskip

\begin{subtable}[t]{1\textwidth}
\scriptsize
\setlength\tabcolsep{6pt}
\begin{center}
\def\arraystretch{1.5}
\begin{tabular}{|| ll | lll | lll | lll ||}
\hline 
\rowcolor{ColorTable}
    &
    & \multicolumn{3}{c|}{Random} 
    & \multicolumn{3}{c|}{Circular} 
    & \multicolumn{3}{c|}{Linear} 
\\
\rowcolor{ColorTable}

 $\alpha$ & Algorithm 
 & $\varepsilon_{\text{r}}$ (deg) & $\varepsilon_{\text{t}}$ (cm) & $\alpha$
 & $\varepsilon_{\text{r}}$ (deg) & $\varepsilon_{\text{t}}$ (cm) & $\alpha$
 & $\varepsilon_{\text{r}}$ (deg) & $\varepsilon_{\text{t}}$ (cm) & $\alpha$
\\ \hline
\multirow{7}{0pt}{\rotatebox{90}{best rotation $\alpha~~~~~~~$}}
& DQOpt
&0.0523$\pm$ $\!\! \tiny\text{0.0162}\atop \!\! \text{0.0156}$&0.1857$\pm$ $\!\! \tiny\text{0.0487}\atop \!\! \text{0.0563}$&0.26
&6.29$\pm$ $\!\! \tiny\text{2.17}\atop \!\! \text{2.02}$&\textbf{42.5}$\pm$ $\!\! \tiny\text{\textbf{8.92}}\atop \!\! \text{\textbf{9.09}}$&0.57
&8.31$\pm$ $\!\! \tiny\text{2.63}\atop \!\! \text{2.42}$&\textbf{45.3}$\pm$ $\!\! \tiny\text{\textbf{8.94}}\atop \!\! \text{\textbf{8.33}}$&0.62
\\& DQ2ndOrd
&0.0523$\pm$ $\!\! \tiny\text{0.0162}\atop \!\! \text{0.0156}$&0.1857$\pm$ $\!\! \tiny\text{0.0487}\atop \!\! \text{0.0563}$&0.26
&6.29$\pm$ $\!\! \tiny\text{2.17}\atop \!\! \text{2.02}$&\textbf{42.5}$\pm$ $\!\! \tiny\text{\textbf{8.92}}\atop \!\! \text{\textbf{9.09}}$&0.57
&8.31$\pm$ $\!\! \tiny\text{2.63}\atop \!\! \text{2.42}$&\textbf{45.3}$\pm$ $\!\! \tiny\text{\textbf{8.94}}\atop \!\! \text{\textbf{8.33}}$&0.62
\\& DQConvRlx
&0.0523$\pm$ $\!\! \tiny\text{0.0162}\atop \!\! \text{0.0156}$&0.1858$\pm$ $\!\! \tiny\text{0.0487}\atop \!\! \text{0.0564}$&0.26
&\textbf{6.25}$\pm$ $\!\! \tiny\text{\textbf{2.22}}\atop \!\! \text{\textbf{1.98}}$&\textbf{42.5}$\pm$ $\!\! \tiny\text{\textbf{8.91}}\atop \!\! \text{\textbf{9.03}}$&0.57
&\textbf{8.25}$\pm$ $\!\! \tiny\text{\textbf{2.66}}\atop \!\! \text{\textbf{2.34}}$&\textbf{45.3}$\pm$ $\!\! \tiny\text{\textbf{8.98}}\atop \!\! \text{\textbf{8.24}}$&0.62
\\& DQ2Steps
&0.0532$\pm$ $\!\! \tiny\text{0.0171}\atop \!\! \text{0.0152}$&0.1846$\pm$ $\!\! \tiny\text{0.0591}\atop \!\! \text{0.0526}$& - 
&9.02$\pm$ $\!\! \tiny\text{3.24}\atop \!\! \text{2.86}$&42.9$\pm$ $\!\! \tiny\text{8.91}\atop \!\! \text{8.90}$& - 
&12.9$\pm$ $\!\! \tiny\text{4.30}\atop \!\! \text{3.72}$&46.2$\pm$ $\!\! \tiny\text{8.47}\atop \!\! \text{8.34}$& -

\\& Dan \cite{Daniilidis1999HandEyeCU}
&0.0524$\pm$ $\!\! \tiny\text{0.0161}\atop \!\! \text{0.0157}$&0.1889$\pm$ $\!\! \tiny\text{0.0516}\atop \!\! \text{0.0548}$&0.26
&17.0$\pm$ $\!\! \tiny\text{7.93}\atop \!\! \text{5.73}$&347.$\pm$ $\!\! \tiny\text{139.}\atop \!\! \text{98.8}$&42.1
&21.9$\pm$ $\!\! \tiny\text{12.1}\atop \!\! \text{7.22}$&499.$\pm$ $\!\! \tiny\text{242.}\atop \!\! \text{139.}$&29.9
\\& ChVecOpt
&\textbf{0.0491}$\pm$ $\!\! \tiny\text{\textbf{0.0170}}\atop \!\! \text{\textbf{0.0140}}$&0.1783$\pm$ $\!\! \tiny\text{0.0521}\atop \!\! \text{0.0504}$&1.04
&6.43$\pm$ $\!\! \tiny\text{2.53}\atop \!\! \text{1.93}$&45.6$\pm$ $\!\! \tiny\text{9.01}\atop \!\! \text{9.04}$&0.57
&8.65$\pm$ $\!\! \tiny\text{2.54}\atop \!\! \text{2.38}$&51.3$\pm$ $\!\! \tiny\text{8.93}\atop \!\! \text{8.32}$&1.00
\\& QuatVecOpt
&0.0500$\pm$ $\!\! \tiny\text{0.0148}\atop \!\! \text{0.0151}$&\textbf{0.1753}$\pm$ $\!\! \tiny\text{\textbf{0.0503}}\atop \!\! \text{\textbf{0.0508}}$&0.26
&6.60$\pm$ $\!\! \tiny\text{2.16}\atop \!\! \text{2.07}$&45.7$\pm$ $\!\! \tiny\text{9.52}\atop \!\! \text{8.19}$&0.26
&8.31$\pm$ $\!\! \tiny\text{2.85}\atop \!\! \text{2.47}$&50.5$\pm$ $\!\! \tiny\text{9.07}\atop \!\! \text{8.55}$&0.34
\\\hline\multirow{7}{0pt}{\rotatebox{90}{best translation $\alpha~~~~~~~$}}
& DQOpt
&0.0527$\pm$ $\!\! \tiny\text{0.0171}\atop \!\! \text{0.0148}$&0.1786$\pm$ $\!\! \tiny\text{0.0555}\atop \!\! \text{0.0526}$&0.24
&6.56$\pm$ $\!\! \tiny\text{2.13}\atop \!\! \text{2.32}$&40.9$\pm$ $\!\! \tiny\text{10.1}\atop \!\! \text{8.18}$&0.62
&\textbf{9.56}$\pm$ $\!\! \tiny\text{\textbf{3.03}}\atop \!\! \text{\textbf{2.90}}$&45.0$\pm$ $\!\! \tiny\text{8.34}\atop \!\! \text{7.71}$&1.14
\\& DQ2ndOrd
&0.0527$\pm$ $\!\! \tiny\text{0.0171}\atop \!\! \text{0.0148}$&0.1786$\pm$ $\!\! \tiny\text{0.0555}\atop \!\! \text{0.0526}$&0.24
&6.56$\pm$ $\!\! \tiny\text{2.13}\atop \!\! \text{2.32}$&40.9$\pm$ $\!\! \tiny\text{10.1}\atop \!\! \text{8.18}$&0.62
&\textbf{9.56}$\pm$ $\!\! \tiny\text{\textbf{3.03}}\atop \!\! \text{\textbf{2.90}}$&45.0$\pm$ $\!\! \tiny\text{8.35}\atop \!\! \text{7.71}$&1.14
\\& DQConvRlx
&0.0527$\pm$ $\!\! \tiny\text{0.0171}\atop \!\! \text{0.0148}$&0.1786$\pm$ $\!\! \tiny\text{0.0554}\atop \!\! \text{0.0525}$&0.24
&\textbf{6.55}$\pm$ $\!\! \tiny\text{\textbf{2.12}}\atop \!\! \text{\textbf{2.40}}$&40.9$\pm$ $\!\! \tiny\text{10.0}\atop \!\! \text{8.21}$&0.62
&\textbf{9.56}$\pm$ $\!\! \tiny\text{\textbf{3.03}}\atop \!\! \text{\textbf{2.85}}$&45.0$\pm$ $\!\! \tiny\text{8.50}\atop \!\! \text{7.51}$&1.14
\\& DQ2Steps
&0.0532$\pm$ $\!\! \tiny\text{0.0171}\atop \!\! \text{0.0152}$&0.1846$\pm$ $\!\! \tiny\text{0.0591}\atop \!\! \text{0.0526}$& - 
&9.02$\pm$ $\!\! \tiny\text{3.24}\atop \!\! \text{2.86}$&42.9$\pm$ $\!\! \tiny\text{8.91}\atop \!\! \text{8.90}$& - 
&12.9$\pm$ $\!\! \tiny\text{4.30}\atop \!\! \text{3.72}$&46.2$\pm$ $\!\! \tiny\text{8.47}\atop \!\! \text{8.34}$& -

\\& Dan \cite{Daniilidis1999HandEyeCU}
&0.0599$\pm$ $\!\! \tiny\text{0.0213}\atop \!\! \text{0.0172}$&0.1809$\pm$ $\!\! \tiny\text{0.0555}\atop \!\! \text{0.0532}$&0.96
&17.0$\pm$ $\!\! \tiny\text{7.93}\atop \!\! \text{5.73}$&347.$\pm$ $\!\! \tiny\text{139.}\atop \!\! \text{98.8}$&42.1
&22.9$\pm$ $\!\! \tiny\text{12.9}\atop \!\! \text{7.52}$&497.$\pm$ $\!\! \tiny\text{263.}\atop \!\! \text{148.}$&21.2
\\& ChVecOpt
&\textbf{0.0506}$\pm$ $\!\! \tiny\text{\textbf{0.0154}}\atop \!\! \text{\textbf{0.0140}}$&\textbf{0.1689}$\pm$ $\!\! \tiny\text{\textbf{0.0558}}\atop \!\! \text{\textbf{0.0480}}$&0.96
&8.77$\pm$ $\!\! \tiny\text{3.47}\atop \!\! \text{2.58}$&30.9$\pm$ $\!\! \tiny\text{11.4}\atop \!\! \text{10.0}$&0.02
&12.6$\pm$ $\!\! \tiny\text{4.53}\atop \!\! \text{3.76}$&\textbf{34.0}$\pm$ $\!\! \tiny\text{\textbf{10.5}}\atop \!\! \text{\textbf{9.34}}$&0.01
\\& QuatVecOpt
&0.0514$\pm$ $\!\! \tiny\text{0.0155}\atop \!\! \text{0.0146}$&0.1719$\pm$ $\!\! \tiny\text{0.0516}\atop \!\! \text{0.0495}$&0.24
&9.15$\pm$ $\!\! \tiny\text{2.95}\atop \!\! \text{2.99}$&\textbf{30.4}$\pm$ $\!\! \tiny\text{\textbf{10.3}}\atop \!\! \text{\textbf{8.93}}$&0.01
&12.4$\pm$ $\!\! \tiny\text{4.09}\atop \!\! \text{3.48}$&34.3$\pm$ $\!\! \tiny\text{10.2}\atop \!\! \text{10.1}$&0.02
\\\hline
\end{tabular}
\end{center}
\end{subtable}
\caption{\small \textbf{ Real and synthetic data experiments}. We show the median and distance to the 25\% and 75\% percentiles of the error distribution of the absolute rotation (geodesic) and translation (Euclidean) errors ($\varepsilon_r$ and $\varepsilon_t$). We separate the results according to different types of motions. The $\alpha$ values are picked to give either the minimal rotation or translation mean error.  We highlight the best median results for each motion type and best rotation/translation $\alpha$. Notice that it does not necessarily indicate significance, where often the different results are very close. \textbf{DQ2Steps} does not depend on $\alpha$ so its estimates are the same for best rotation/translation $\alpha$. }
\label{tb:real_data_exp}
\end{table*}

\subsection{Verifying optimality}
In order to verify that our algorithm finds the optimal solution we compare its cost with the cost of a nonlinear optimization implementation (using scipy's least\_squares), designed to minimize the same cost function. When running the nonlinear optimizer we initialized it both with our solution and randomly to allow a search for a different minimum in case our algorithm ended up in a local minimum.
In order for our algorithm to be optimal, the nonlinear optimizer should not change the solution when initialized with our solution, and might find higher cost solutions when initialized randomly. The results are summarized in table \ref{tb:optimality}, where our algorithm always returns a lower cost up to numerical precision. We further compare with our approximate methods to show how close they get to the global minimum.

\subsection{Timing}
We compare the run time of our optimal solution implementation with the approximate solutions and \textbf{Dan} \cite{Daniilidis1999HandEyeCU}, which serves as a benchmark, as it is well known to be very efficient.
The results are shown in lower table \ref{tb:optimality}, ordered by the mean run time. Our optimal algorithm is found to be $\times 3$ slower than \textbf{Dan}. Two of the approximate algorithms are a bit faster, while approximating very well the optimal solution, see upper table \ref{tb:optimality}. The nonlinear optimization algorithms are much slower, but we do not report their performance since we did not optimize their implementation.
\section{Conclusions}
\label{sec:Conclusions}

In this paper we introduced a novel approach to find an optimal solution to the hand-eye calibration problem, based on a dual-quaternions formulation.
We showed that our algorithm is guaranteed to find the global minimum by only using a 1D line search of a convex function.
This provides a big advantage over nonlinear optimization methods which are less efficient and are sensitive to local minima.

We further introduced a hierarchy of efficient analytic approximate solutions which provide remarkably close approximations to the optimal solution.
We showed that our optimal solution is very efficient being only 3 times slower than the  analytic solution by Daniilidis \cite{Daniilidis1999HandEyeCU}, and that some of our approximation are faster while providing dramatically better estimations for ill-posed cases (comparable with the optimal solution).

We compared the performance of our algorithms with the well known analytic solution by Daniilidis and nonlinear optimization of other common formulations of the problem with different cost functions, using real and synthetic noisy datasets.
We find that all the methods which provide an optimal solution are comparable in terms of accuracy, see table \ref{tb:real_data_exp}.
However, for the real data we find that our methods consistently obtain slightly lower median errors in terms of both rotations and translations.
The two-steps method degrades in accuracy for almost planar motions due to the ill-posedness of the rotation equation.
Daniilidis's algorithm which does not solve an optimization problem shows a degradation in accuracy in the same scenarios.

We also introduced a prior to our formulation which does not change the efficiency nor the optimality of our solution.
The prior term can help for having maximum a posterior probability (MAP) analysis, or to regularize the estimation for ill-posed scenarios.
Other generalizations, such as adding weights or a covariance, can be trivially incorporated to our formulation.

Generally we find that optimal methods are more robust to almost planar motions, however the best performing method will depend on the trajectory and noise in the given problem which dictates the residuals distribution.
Our method has the advantage of providing good results comparable with other optimal methods, while being very efficient.

\section*{Acknowledgments}

This research was done as part of a computer vision research project at Zenuity.  
The authors gratefully acknowledge funding and support from Zenuity.
We would also like to thank Ernesto Nunes for useful discussions.

{\small
\bibliographystyle{unsrt}
\bibliography{references}
}


\clearpage
\newpage

\section*{Supplemental Material}\label{sec:supplemental_material}

\appendix
\section{Deriving the second order approximations (DQ2ndOrd)}\label{app:second_order}
  In this appendix we give a detailed derivation of the expanded solution in small $\mu$ around the relaxed solution (\ref{eq:conv_relax}).
  We first define an expansion of the quaternion solution (eigenvector) and the smallest eigenvalue of (\ref{eq:ev1})
 \begin{align}
 q = \sum_{k=0}^\infty \tilde q_k \mu^k, \quad
 \lambda = \sum_{k=0}^\infty \tilde \lambda_k \mu^k,
 \end{align}	
 such that (\ref{eq:ev1}) is solved order by order in $\mu$, and the constraint $q^2=1$ is satisfied.
 We can fix $\mu$ by using the second constraint $q\cdot q' = 0$.
 We use $Z_0 q_a = \lambda_a q_a$, where we order the solutions so that $\lambda_0 < \lambda_1<  \lambda_2<\lambda_3$, as a basis for the quaternions at every order,so that
  \begin{align}
 \tilde q_k  = \sum_{a=0}^3 c_{k, a} q_a,
 \end{align}
 where $c_{k, a}$ are coefficients to be fixed.
 Plugging these into \eqref{eq:ev1} and the constraint equations we find the following recursion relations (for $k\geq 1$)
 \begin{align}
& c_{k, 0}  = - \frac{1}{2}\sum_{n=1}^{k-1} \tilde q_{k-n}\cdot \tilde q_n,\nonumber\\
 & \tilde \lambda_k = q_0 (Z_1 \tilde q_{k-1} - Z_2 \tilde q_{k-1}) - \sum_{l=1}^{k-1}\tilde \lambda_{k-l} c_{l, 0},\nonumber\\
& c_{k, a} = \frac{q_a(Z_1 \tilde q_{k-1}-Z_2 \tilde q_{k-2}) - \sum_{l=1}^{k-1}\tilde \lambda_{k-l}c_{l, a}}{\lambda_0 - \lambda_a},
 \end{align}
 with $c_{0, 0}=1$, $c_{0, a} = 0$, $\tilde q_{k<0} = 0$, and $\tilde \lambda_0 = \lambda_0$.
 Thus we can easily compute $\lambda(\mu)$ to arbitrary order in $\mu$, and then solve $\frac{d \lambda}{d \mu}|_{\mu^\ast} = 0$.
 Since this is a polynomial equation in $\mu$, we can solve it analytically for low orders.
 In practice we find that the second order approximation yields extremely good results.
 In that case we have 
   \begin{align}
   \mu^\ast_{(2)} = \frac{q_0 Z_1 q_0/2}{q_0 Z_2 q_0 - \sum_{a=1}^3\frac{(q_a Z_1 q_0)^2}{\lambda_0 - \lambda_a}}.
   \end{align}
  After getting $q_{(2)}$ ($q$ to second order in $\mu$) using $\mu^\ast_{(2)}$ and the expansion above and normalizing it, we get the corresponding $q'$ using (\ref{eq:qp_sol}).
  
In the main text we described another possible expansion in the dimensionless parameter $\lambda$.
As explained in that sec. \ref{subsec:2ndOrd}, we expand again around the relaxed solution (and not around $\lambda=0$), so $\lambda = \lambda_0 + \Delta \lambda$ and the expansion is in $\Delta \lambda$, and $\lambda_0$ corresponds to the smallest eigenvalue of $Z_0$.
 Carrying the expansion to second order leads to
 \begin{align}
 \tilde \mu_0 =& 0, \quad, \tilde \mu_1 = \frac{1}{q_0 Z_1 q_0},\quad\tilde \mu_2 = -\frac{\tilde \mu_1 q_0 Z_1 \tilde q_1 + \tilde \mu_1^2 q_0 Z_2 q_0}{q_0 Z_1 q_0},\nonumber\\
 \tilde q_1 =& \sum_{a=1}^3\frac{\tilde \mu_1 q_a Z_1 q_0}{\lambda_0 - \lambda_a} q_a,\\
 \tilde q_2 =& \sum_{a=1}^3\frac{q_a Z_1 \left(\tilde \mu_1 \tilde q_1 +   \tilde \mu_2 q_0\right) + \tilde \mu_1^2 q_a Z_2 q_0 - q_a \tilde q_1}{\lambda_0 - \lambda_a} q_a
 -\frac{\tilde q_1^2}{2}.\nonumber
 \end{align}
 Using these expressions we can compute the constraint to second order
 \begin{align}
 0 =& q(\mu Z_2 + \frac{Z_1}{2})q = 
 \frac{1}{2}q_0 Z_1 q_0 + \Delta \lambda q_0 \left(Z_1 \tilde q_1 + \tilde \mu_1 Z_2 q_0\right)\nonumber\\
 &+\Delta \lambda^2\left(q_0 Z_1 \tilde q_2 + \frac{1}{2}\tilde q_1 Z_1 \tilde q_1
 + q_0 Z_2 (2\tilde \mu_1 \tilde q_1
 + \tilde \mu_2 q_0)\right)\nonumber\\
 \equiv &~
c_0 + \Delta \lambda c_1 + \Delta \lambda^2 c_2,
 \end{align}
 solve for $\Delta \lambda = (-c_1 + \sqrt{c_1^2 - 4c_0 c_2})/2$ and plug for $\tilde \mu = \Delta \lambda(\tilde \mu_1 + \tilde \mu_2 \Delta \lambda)$, and compute the corresponding solution for $q$ and then the unique $q'$ using (\ref{eq:qp_sol}).

\section{Optimal polynomial solution (DQOptPoly)}\label{sec:polystuff}
In sec. \ref{sec:prorp_of_problem} we explained that the optimal solution to the hand-eye calibration problem can be found in a closed form by solving a 28-th degree univariant polynomial. 
However, the explicit polynomial coefficients in terms of the matrix entries of (\ref{eq:ev1}) are quite complicated, consisting of thousands of terms which makes this approach less attractive for practical use.
Therefore, we presented an efficient way to arrive at the solution without directly solving the polynomial. In this appendix we present two more approaches to arrive at the optimal solution which are also based on e 1D line search, but of a very different function. These methods yield the same solution, but are a bit slower than the one presented in sec. \ref{sec:prorp_of_problem}. However, since they are conceptually different we find it useful to describe them in this section.

Throughout this section we assume that the 25 polynomial coefficient of the algebraic curve \eqref{eq:alg_crv} have been extracted from the matrices $Z_i$ (their explicit form is quite lengthy and not very illuminating, so we do not state it here. In any case it is straight forward to extract it from \eqref{eq:alg_crv}.).

An alternative approach is to find the $\lambda$ root of the resultant for the $\mu$ polynomial with $\lambda$ dependent coefficients numerically. Namely finding the root of a function that return the resultant given the 25 coefficients and $\lambda$ which is easy to compute (in contrast to the explicit coefficients of the 28-degree polynomial). Alternatively we can repeat the same procedure for the $\lambda$ polynomial with $\mu$ dependent coefficients.
The root search can be bounded, see for example appendix \ref{app:bounds}. 

Another approach to arrive at the solution is to 
count the number of real roots of the 8-th degree $\mu$ polynomial with $\lambda$ dependent coefficients. For $\lambda=0$ we have eight real roots, this number remains as we increase $\lambda$, and jumps to six exactly as we pass the optimal solution ($\lambda$ corresponding to the optimal solution). The analytic expression for the number of roots is known in terms of Sturm's theorem.
To compute the number of real roots one needs to compute the Sturm sequence defined by $P_0(\mu)$ which is the polynomial, $P_1 = \dot P_0$, $P_2 = -\text{rem}(P_0, P_1)$ and so on up to $P_8 = -\text{rem}(P_6, P_7)$ which is a constant. Then, plugging $\pm \infty$ and computing the difference in the number sign changes between the elements of the sequence gives the number of real roots. 
Thus, we can set the polynomial at $\lambda=0$ such that the sequences signs at $-\infty$ and $+\infty$ are $(-,+,-,+,-,+,-,+,-)$ and $(+,+,+,+,+,+,+,+,+)$ respectively, which in turn implies there are eight real solutions.
A drop from eight to six real solutions occurs when the first ($P_0$) or the last ($P_8$) term in the sequence changes sign, however, the sign of $P_0(\pm\infty)$ does not depend on $\lambda$, so we are only interested in the $\lambda$ dependence of $P_8$, and more concretely, on the minimal positive root of $P_8$ with respect to $\lambda$ which corresponds to our solution.

As before, the explicit analytic expression is quite involved and in practice one can use 1D numerical root finder for the explicit real roots counting number function to detect the jump from eight to six. 
  
 \section{Analytic bounds on the parameter space}\label{app:bounds}
 We can analytically bound the Lagrange multiplier variables as follows.
 First, we are only interested in the $\lambda \geq 0$ region, since as mentioned in the main text $\lambda$ equals the cost function, which is non-negative by construction, and so any $\lambda$ corresponding to a solution must be non-negative.
 
 Next, we can also bound $\mu$ by noticing that we can rewrite (\ref{eq:mu_eq}) as $\mu = \frac{(Lq)^T K (Lq)}{|Lq|^2}$, where $M^{-1} = L^T L$ is the Cholesky decomposition and $K\equiv \frac{1}{2}(LW^TL^{-1}+L^{-T}WL^T)$. From this expression we get a bound on $\mu$ in terms of the largest and smallest eigenvalues of $K$.

 \section{Experiments}\label{app:datasets}

We validate out algorithms on a combination of real world and synthetic datasets. Each dataset consists of a set of corresponding noisy measurements $\Delta \tilde P_1^i$, $\Delta \tilde P_2^i$ of relative poses from two rigidly attached sensors moving in 3D. 

From each dataset draw $N=1000$ samples of $n=100$ corresponding relative poses (with replacement between the samples) from which we generate a histogram of translation and rotation errors for each algorithm. In order to reduce the impact of temporal correlations within the  samples the $n$ relative poses are drawn randomly from entire trajectory and are thus not necessarily consecutive in time. 

In order to account for the effect of the weighting parameter we repeat the experiments for 100 values of $\alpha$ evenly distributed on a logarithmic scale between $10^{-2}$ and $10^{1.7}$ (we found that no algorithm performs well outside this range for the used datasets). In Table \ref{tb:real_data_exp} we display the median rotation and translation error together with the 25\% and 75\% percentiles for the $\alpha$ that gives the lowest median rotation and translation error respectively for each dataset and algorithm.

\subsection{Real data experiments}

\begin{figure}[t]
  \begin{center}
   \includegraphics[width=\linewidth]{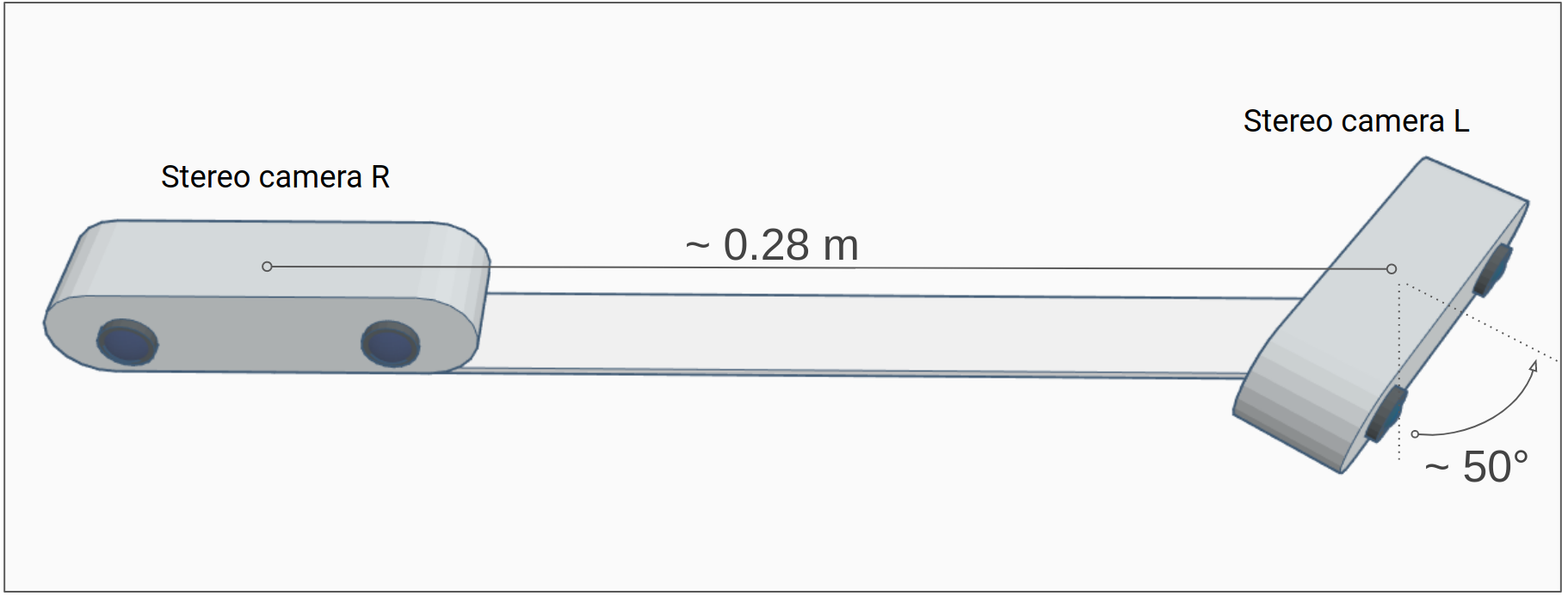}
   \caption{\small{Our experimental setup. Two rigidly attached stereo cameras.}}
   \label{fig:hand_eye_setup_real}
  \end{center}
\end{figure}

\begin{figure*}[htpb]
\centering
\includegraphics[width=0.96\linewidth]{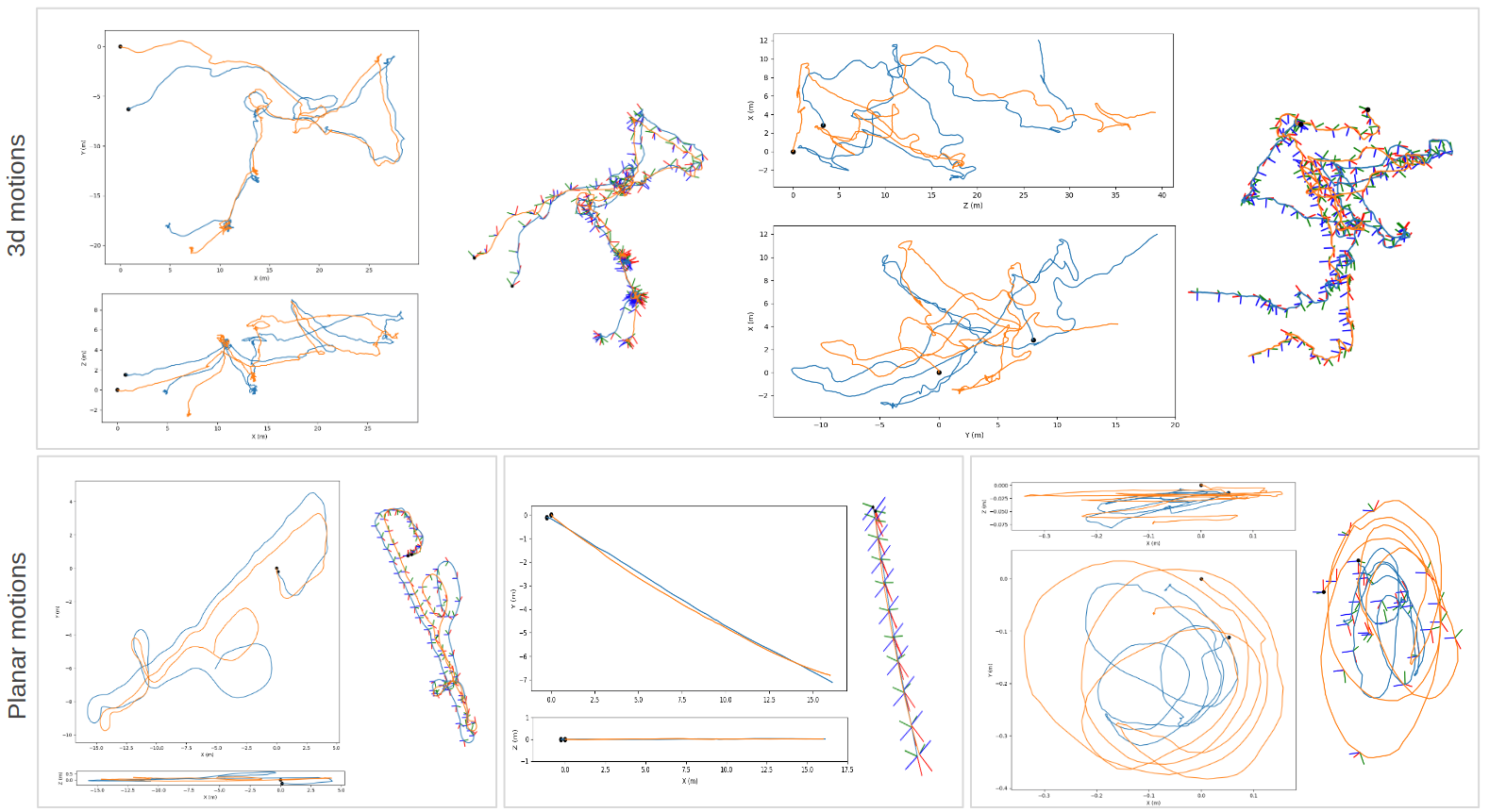} 
\caption{\small The two estimated camera trajectories from the real data experiments shown in blue and orange respectively. We show both 2d projections and a 3d plot for each dataset. The data used here has been filtered by taking away relative poses with large errors in rotation or translation \wrt the ground truth calibration, which we consider as outliers where the SLAM algorithm fails. The upper plots show the general 3d motion trajectories and the lower plots show the Planar, Linear and Rotation trajectories from left to right.}
\label{fig:real_exp_traj}
\end{figure*}

First, we validate our algorithms on four datasets generated by a multi-camera tracking system. Two ZED stereo cameras\footnote{ZED stereo camera: https://www.stereolabs.com/zed/} were rigidly attached at circa 28 cm from each other at an angle of $50^\circ$, maintaining partially overlapping fields of view as shown in fig.~\ref{fig:hand_eye_setup_real}. The multi-camera setup was manually moved on a large area (circa 130 $\text{m}^2$) following several 3D trajectories shown in fig.~\ref{fig:real_exp_traj}.

Two SLAM components running on the stereo camera provided 6 \dof poses at 5 Hz ($C_i$ and $H_i$ in fig.~ \ref{fig:hand_eye_setup}). We ran the two visual tracking systems at limited resolution settings (1280x720 pixels) independently from each sensor, \ie no shared features and different reference frames as shown in fig.~ \ref{fig:hand_eye_setup}).
Loop closure and mapping was also disabled to avoid drift and scale correction over long, redundant trajectories.
The true extrinsic $X$ was estimated offline, by placing a large fiducial marker in the overlapping FOV of the cameras. Multiple high resolution pictures (3840x2160 pixels), taken from the two cameras, were used to determine the position of the marker \wrt the camera optical centers. The sensor manufacturer's intrinsics and extrinsics were then used to compute the relative pose between the two stereo camera centers, corresponding to the tracked points. The estimated relative pose ($X$ in fig.~ \ref{fig:hand_eye_setup}) resulting from this calibration procedure had translation: [-0.7,  28.1, -0.1] cm and axis-angles: [2.35,  -0.92, -48.93] deg. 
 
In order to reduce the impact of outliers from the tracking algorithms we pre-filtered the data by discarding any datapoints where either of the cameras has moved more than 10 cm or $11.5^\circ$ between two consecutive frames, as that indicates tracking error. These thresholds were empirically determined considering the speed with which the two cameras were moved during the experiments. In order to reduce the effect of synchronization issues we also removed any datapoints where the timestamps from the cameras differ by more than 0.1 seconds, the poses that are left differ in timestamp by roughly 0.05 seconds.

In our experiments we recorded four types of motion (see fig.~\ref{fig:real_exp_traj}). First we recorded a general 3d motion where the cameras device was moved freely by hand in a large room. The other types of motion are approximately planar, where we placed the device on a cart which was moved by hand across the room. In this way we created three more trajectories: a general planar motion, a linear motion (where the cart was moved only forward) and a circular motion (where the cart was rotated). 

\subsection{Synthetic datasets}
We also test our algorithm on three different scenarios of synthetically generated data, where the ground-truth is known exactly. For each scenario we generate a set of $N$ corresponding relative poses $\Delta P_1^i$, $\Delta P_2^i$ with 
\begin{align}
    \Delta P_2^i = X  \circ \Delta P_1^i \circ X^{-1}.
\end{align} 
We simulate noisy measurements $\Delta \tilde P_k^i$ of $\Delta P_k^i$ by applying small $SE(3)$-perturbations:
\begin{align}
    \Delta \tilde P_k^i = \Delta P_k^i \circ [\delta R_k^i(\sigma_r) | \delta t_k^i(\sigma_t)],
\end{align}
where $\delta R_k^i(\sigma_r)$ is generated by drawing its rotation axis from a uniform $SO(3)$ distribution \cite{Arvo:1992:FRR:130745.130767} and its angle from a normal distribution\footnote{For simplicity we use a normal distribution for the angle, though it would be more appropriate to use the more complicated wrapped normal distribution that takes the periodicity into account.} with variance $\sigma_t^2$. $\delta t_k^i(\sigma_t)$ is generated by drawing its elements from a normal distribution with variance $\sigma_t^2$. In Table \ref{tb:real_data_exp} we have used $\sigma_r=0.57^\circ$ and $\sigma_t=0.01$ m for the pose noise.

We consider the three following scenarios:

\noindent
- \textbf{Random} The relative poses $\Delta P_1^i$ of the first trajectory are generated by drawing the rotational part from a uniform distribution on $SO(3)$ \cite{Arvo:1992:FRR:130745.130767}, and the translational part from $U(0, 1)^3$.

\noindent
- \textbf{Line} Constant velocity along the x-axis from $x=0$ to $x=2$ and constant orientation.

\noindent
- \textbf{Circle} One revolution of a circle with radius 2.

\begin{figure*}[bhtp]
\centering
\includegraphics[width=0.96\linewidth]{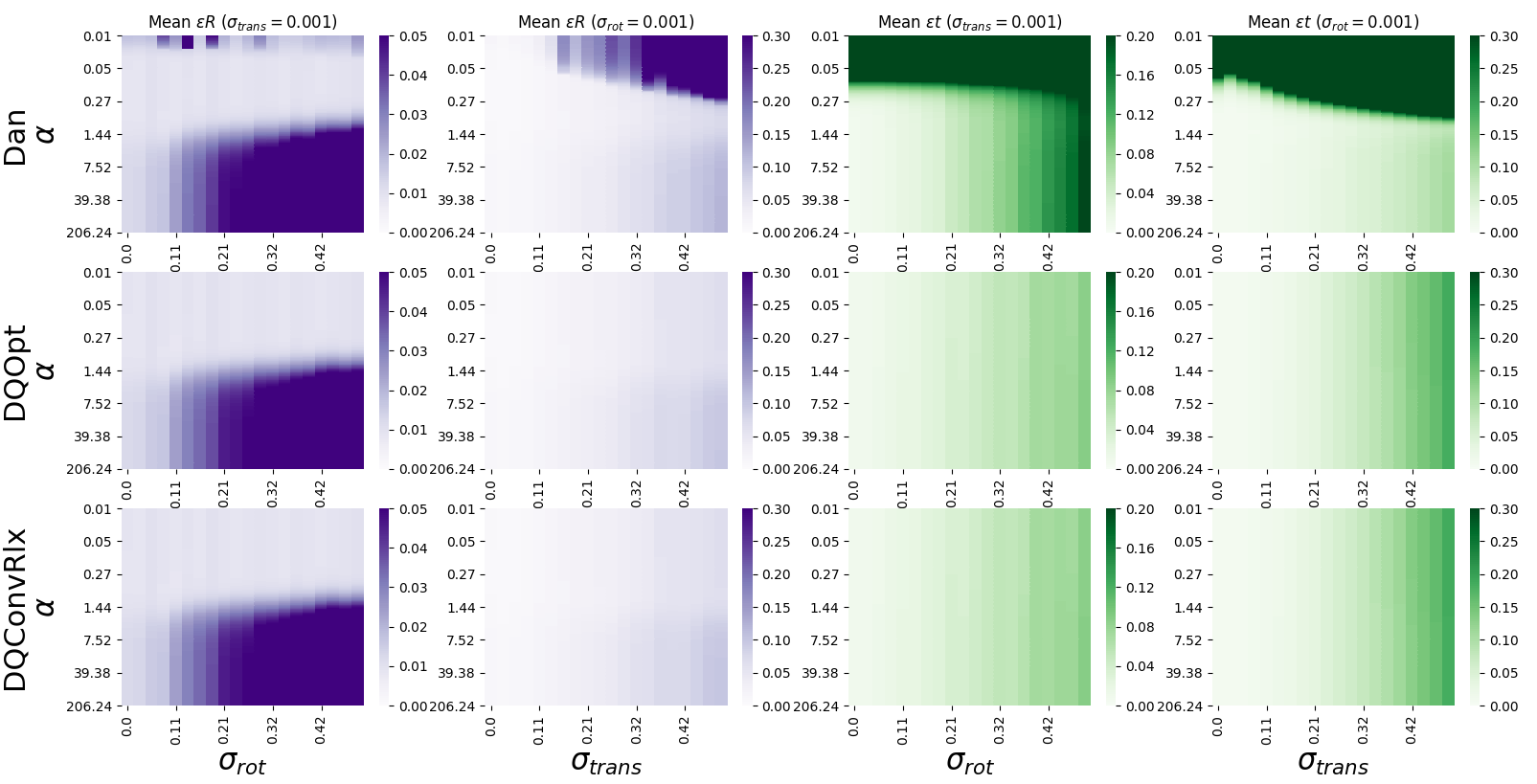} 
\caption{\small Effect of $\alpha$ on the translation and rotation errors for \textbf{Dan}, \textbf{DQOpt} and \textbf{DQConvRlx} for the Random synthetic scenario with various noise levels. The first two columns (in purple) show the mean rotation errors and the last two (in green) show the mean translation errors, each averaged over 60 samples. We vary the noise level on the translation or rotation component respectively, while keeping the noise on the remaining component fixed.}
\label{fig:experiments_sssynt}
\end{figure*}

In order to break the symmetry for the line and circle scenarios we apply a perturbation with  $\sigma_r=0.57^\circ$ and $\sigma_t=0.01$ m to each $\Delta P_1^i$ before computing $\Delta P_2^i$, simulating a slight jitter in the trajectory. For the Random scenario we plot the algorithm performance as a function of $\alpha$ at various noise levels in fig.~ \ref{fig:experiments_sssynt}. 

\section{Algorithms summary}

In this appendix we summarize in pseudo-code our algorithms which appear in the main text.
We describe the algorithms after processing the data, namely, given the matrices $S, W$ and $M$ in (\ref{eq:cost_function}). The $Z_i$ matrices are defined in (\ref{eq:ev1}). Throughout this section, by smallest eigenvector we mean the eigenvector corresponding to the smallest eigenvalue.
\begin{algorithm}
\small
\KwData{$S$, $W$, $M$}
\KwResult{$q_\ast$, ${q'}_\ast$}
  \caption{\small 1D line search (\textbf{DQOpt})}\label{alg:1dlinesearch}
  \begin{algorithmic}[1]
    \Function{$f$}{$\mu$}
        \State{$q := \text{smallest Eigenvector}(Z_0 + \mu Z_1 - \mu^2 Z_2)$}
        \State\Return $q^T \left(\mu Z_2 - \frac{1}{2}Z_1\right) q$
    \EndFunction
    \State Use root finding procedure to solve, $f(\mu_\ast) = 0$
    \State $q_\ast := \text{smallest Eigenvector}(Z_0 + \mu_\ast Z_1 - \mu_\ast^2 Z_2)$
    \State ${q'}_\ast := Z_2(\mu_\ast - W^T)q_\ast$
  \end{algorithmic}
\end{algorithm}
\setlength{\intextsep}{7pt}
\begin{algorithm}
\small
\KwData{$W$, $M$}
\KwResult{$q_\ast$, ${q'}_\ast$}
  \caption{\small 2 steps algorithm (\textbf{DQ2Steps})}\label{alg:2steps}
  \begin{algorithmic}[1]
    \State $q_\ast := \text{smallest Eigenvector} (M) $
    \State ${q'}_\ast := Z_2\left(\frac{1}{2}\frac{q_\ast^T Z_1 q_\ast}{q_\ast^T Z_2 q_\ast} - W^T\right)q_\ast$
  \end{algorithmic}
\end{algorithm}

\begin{algorithm}
\small
\KwData{$S$, $W$, $M$}
\KwResult{$q_\ast$, ${q'}_\ast$}
  \caption{\small Convex relaxation algorithm (\textbf{DQConvRlx})}\label{alg:cnvxrlx}
  \begin{algorithmic}[1]
    \State $q_\ast := \text{smallest Eigenvector} (Z_0) $
    \State ${q'}_\ast := Z_2\left(\frac{1}{2}\frac{q_\ast^T Z_1 q_\ast}{q_\ast^T Z_2 q_\ast} - W^T\right)q_\ast$
  \end{algorithmic}
\end{algorithm}

\begin{algorithm}
\small
\KwData{$S$, $W$, $M$}
\KwResult{$q_\ast$, ${q'}_\ast$}
  \caption{\small second order algorithm (\textbf{DQ2ndOrd})}\label{alg:cnvxrlx}
  \begin{algorithmic}[1]
    \State Solve $Z_0 q_i = \lambda_i q_i$ for $q_i$ and $\lambda_i$, $i=0,...,3$
    \State Compute $q_{(2)}$ using \eqref{eq:sec_ord}
    \State $q_\ast = q_{(2)} / q_{(2)}^2$
    \State ${q'}_\ast := Z_2\left(\frac{1}{2}\frac{q_\ast^T Z_1 q_\ast}{q_\ast^T Z_2 q_\ast} - W^T\right)q_\ast$
  \end{algorithmic}
\end{algorithm}

\begin{algorithm}
\small
\SetAlgoLined
\KwData{$S$, $W$, $M$, $\epsilon$}
\KwResult{$q_\ast$, ${q'}_\ast$}
$\mu := 0$\;
\While{$\Delta > \epsilon$}{
  $q := \text{smallest Eigenvector}(Z_0 + \mu Z_1 - \mu^2 Z_2)$\;
  $\mu_{\text{new}} =  \frac{1}{2}\frac{q(\mu)^T Z_1 q(\mu)}{q(\mu)^T Z_2 q(\mu)}$\;
  $\Delta := |\mu - \mu_{\text{new}}|$\;
  $\mu := \mu_{\text{new}}$\;
  }
$q_\ast := q$\\
${q'}_\ast := Z_2(\mu_\ast - W^T)q_\ast$\\ 
 \caption{\small Iterative algorithm (\textbf{DQItr})}
\end{algorithm}


\end{document}